\documentclass{article}

\usepackage{iclr2026_conference,times}

\usepackage{microtype}
\usepackage{amsmath}
\usepackage{amssymb}
\usepackage{amsthm}
\usepackage{mathtools}
\usepackage{multirow}
\usepackage{booktabs}
\usepackage{tabularx}
\usepackage{booktabs}
\usepackage{graphicx}
\usepackage{xcolor}
\usepackage{enumitem}
\usepackage{listings}
\usepackage[most]{tcolorbox}
\usepackage{float}
\usepackage{tikz}
\usetikzlibrary{arrows.meta}
\usepackage{url}
\usepackage{hyperref}

\definecolor{draftred}{RGB}{150,30,30}
\definecolor{draftgray}{RGB}{245,245,245}
\definecolor{opnavy}{HTML}{173B5E}
\definecolor{opteal}{HTML}{168A81}
\definecolor{opgold}{HTML}{E6A23C}

\newcommand{\method}{\ensuremath{\mathrm{OP}^{2}\mathrm{SD}}}
\newcommand{\targetopsd}{\textsc{OPSD}}

\newcommand{\junpei}[1]{} 

\theoremstyle{remark}

\title{Privileged Solutions or Context-Induced Teacher Behavior? Dissecting On-Policy Self-Distillation}

\author{%
  Yuki Ichihara\textsuperscript{1} \quad
  Naoto Iwase\textsuperscript{2} \quad
  Mohammad Atif Quamar\textsuperscript{1} \quad
  Junpei Komiyama\textsuperscript{1,3} \\[2pt]
    \textsuperscript{1}Mohamed bin Zayed University of Artificial Intelligence \quad
  \textsuperscript{2}Nagoya University \quad
  \textsuperscript{3}RIKEN AIP \\[2pt]
    \texttt{\{yuki.ichihara, mohammad.atif\}@mbzuai.ac.ae} \\
  \normalfont\small\texttt{naoto@iwase.dev} \quad
  \texttt{junpei@komiyama.info} \\[2pt]
}

\iclrfinalcopy
\begin{document}

\maketitle

\begin{abstract}
On-Policy Self-Distillation (OPSD) is commonly interpreted as the transfer of privileged information: a teacher observes the verified solution to the target problem and supervises the student's trajectory. However, this interpretation conflates two effects. The reference solution not only reveals the answer to the current instance but also changes the context under which the teacher provides token-level supervision.
We investigate the role of target-specific privilege with \method{}
(\emph{\textbf{O}n-\textbf{P}olicy \textbf{S}elf-\textbf{D}istillation from Other \textbf{P}roblems}), which replaces the
paired reference with a problem and solution from a different example, while preserving the student rollout, teacher, and
distillation objective. Across three models and three mathematics benchmarks, \method{} improves over the base model,
remains competitive with OPSD.
The success of \method{} implies that OPSD gains do not necessarily come from access to the reference solution, and that the teacher's context-induced behavior is an important factor.
Our implementation is available at \url{https://github.com/MBZUAI-reasoninglab/OP2SD}.
\end{abstract}

\definecolor{opnavy}{HTML}{173B5E}
\definecolor{opteal}{HTML}{168A81}
\definecolor{opgold}{HTML}{E6A23C}
\definecolor{opred}{HTML}{C64B4B}
\definecolor{opink}{HTML}{172635}
\definecolor{opmist}{HTML}{F3F6F8}

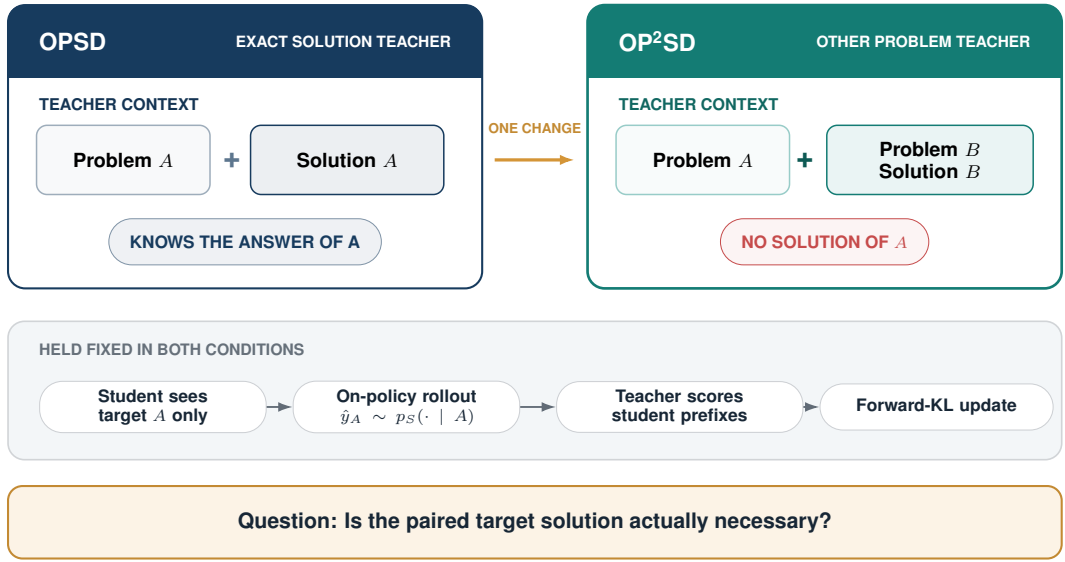
\begin{figure}[h]
\centering
\resizebox{\textwidth}{!}{%
\begin{tikzpicture}[x=1mm,y=1mm]
  \tikzset{
    card/.style={
      rounded corners=2.6mm,
      line width=0.9pt,
      fill=white
    },
    mini/.style={
      rounded corners=1.6mm,
      line width=0.65pt,
      minimum height=10.5mm,
      align=center,
      font=\sffamily\fontsize{8.4}{9.6}\selectfont
    },
    pill/.style={
      rounded corners=3.5mm,
      minimum height=7mm,
      align=center,
      inner xsep=3.2mm,
      font=\sffamily\bfseries\fontsize{7.4}{8.4}\selectfont
    },
    flow/.style={
      -{Latex[length=2.2mm,width=1.5mm]},
      line width=0.8pt,
      draw=opink!68
    }
  }


  \draw[card,draw=opnavy] (-80,-14) rectangle (-8,-57);
  \fill[opnavy,rounded corners=2.3mm] (-80,-14) rectangle (-8,-25);
  \fill[opnavy] (-80,-22.7) rectangle (-8,-25);

  \node[
    anchor=west,
    text=white,
    font=\sffamily\bfseries\fontsize{10.2}{11.5}\selectfont
  ] at (-76.5,-19.5) {OPSD};
  \node[
    anchor=east,
    text=white!80,
    font=\sffamily\bfseries\fontsize{6.7}{7.6}\selectfont
  ] at (-11.5,-19.5) {EXACT SOLUTION TEACHER};

  \node[
    anchor=west,
    text=opnavy!88!black,
    font=\sffamily\bfseries\fontsize{7.0}{8.0}\selectfont
  ] at (-76.5,-29) {TEACHER CONTEXT};

  \node[
    mini,
    draw=opnavy!42,
    fill=opnavy!3,
    text width=24mm
  ] (left-target) at (-62.5,-37.5)
  {\textbf{Problem \(A\)}};

  \node[
    text=opnavy!70,
    font=\sffamily\bfseries\fontsize{13}{13}\selectfont
  ] at (-46,-37.5) {+};

  \node[
    mini,
    draw=opnavy,
    fill=opnavy!8,
    text width=27mm
  ] (left-solution) at (-28.5,-37.5)
  {\textbf{Solution \(A\)}};

  \node[
    pill,
    anchor=south,
    draw=opnavy!55,
    fill=opnavy!7,
    text=opnavy
  ] at (-44,-53.5) {KNOWS THE ANSWER OF A};

  \draw[card,draw=opteal!90!black,line width=1.15pt] (8,-14) rectangle (80,-57);
  \fill[opteal!90!black,rounded corners=2.3mm] (8,-14) rectangle (80,-25);
  \fill[opteal!90!black] (8,-22.7) rectangle (80,-25);

  \node[
    anchor=west,
    text=white,
    font=\sffamily\bfseries\fontsize{10.2}{11.5}\selectfont
  ] at (11.5,-19.5) {OP\textsuperscript{2}SD};
  \node[
    anchor=east,
    text=white!82,
    font=\sffamily\bfseries\fontsize{6.7}{7.6}\selectfont
  ] at (76.5,-19.5) {OTHER PROBLEM TEACHER};

  \node[
    anchor=west,
    text=opteal!75!black,
    font=\sffamily\bfseries\fontsize{7.0}{8.0}\selectfont
  ] at (11.5,-29) {TEACHER CONTEXT};

  \node[
    mini,
    draw=opteal!44,
    fill=opteal!3,
    text width=24mm
  ] (right-target) at (25.5,-37.5)
  {\textbf{Problem \(A\)}};

  \node[
    text=opteal!65!black,
    font=\sffamily\bfseries\fontsize{13}{13}\selectfont
  ] at (41,-37.5) {+};

  \node[
    mini,
    draw=opteal!90!black,
    fill=opteal!9,
    text width=29mm
  ] (right-example) at (60,-37.5)
  {\textbf{Problem \(B\)\\Solution \(B\)}
};

  \node[
    pill,
    anchor=south,
    draw=opred,
    fill=opred!6,
    text=opred
  ] at (44,-53.5) {NO SOLUTION OF \(A\)};

  \draw[
    -{Latex[length=2.3mm,width=1.6mm]},
    draw=opgold!92!black,
    line width=1.4pt
  ] (-6.1,-37.5) -- (6.1,-37.5);
  \node[
    text=opgold!85!black,
    align=center,
    font=\sffamily\bfseries\fontsize{5.9}{6.7}\selectfont
  ] at (0,-32.6) {ONE CHANGE};

  \draw[
    rounded corners=2.5mm,
    draw=opink!20,
    fill=opmist,
    line width=0.7pt
  ] (-80,-62) rectangle (80,-83);

  \node[
    anchor=west,
    text=opink!66,
    font=\sffamily\bfseries\fontsize{6.8}{7.8}\selectfont
  ] at (-76.5,-66.2) {HELD FIXED IN BOTH CONDITIONS};

  \node[
    pill,
    draw=opink!22,
    fill=white,
    text=opink,
    text width=28mm
  ] (student) at (-58,-75) {Student sees\\target \(A\) only};

  \node[
    pill,
    draw=opink!22,
    fill=white,
    text=opink,
    text width=28mm
  ] (rollout) at (-19.5,-75) {On-policy rollout\\\(\hat y_A\sim p_S(\cdot\mid A)\)};

  \node[
    pill,
    draw=opink!22,
    fill=white,
    text=opink,
    text width=31mm
  ] (score) at (22,-75) {Teacher scores\\student prefixes};

  \node[
    pill,
    draw=opink!22,
    fill=white,
    text=opink,
    text width=29mm
  ] (update) at (61,-75) {Forward-KL update};

  \draw[flow] (student.east) -- (rollout.west);
  \draw[flow] (rollout.east) -- (score.west);
  \draw[flow] (score.east) -- (update.west);

  \draw[
    rounded corners=2.2mm,
    draw=opgold!85!black,
    fill=opgold!13,
    line width=0.9pt
  ] (-80,-87) rectangle (80,-98);

  \node[
    text=opink,
    font=\sffamily\bfseries\fontsize{9.2}{10.5}\selectfont
  ] at (0,-92.5)
  {Question: Is the paired target solution actually necessary?};
\end{tikzpicture}%
}
\caption{\targetopsd{} and \method{} share the student prompt, on-policy
rollout, frozen self-teacher fixed to the base model, and loss. The only difference is the
teacher-only context: \targetopsd{} supplies the verified solution to the target problem \(A\), whereas \method{} supplies a worked solution to a different problem \(B\neq A\). In this way, \method{} withholds the target solution while retaining the teacher-only worked-solution context.}
\label{fig:teacher-context-swap}
\end{figure}

\section{Introduction}
On-policy distillation (OPD) \citep{gu2024minillm,agarwal2024onpolicy} has recently emerged as a practical approach to distilling autoregressive language models. Unlike distillation on a fixed dataset of teacher-generated outputs, OPD samples trajectories from the current student and obtains token-level supervision from the teacher along the resulting student-generated prefixes. \citet{agarwal2024onpolicy} demonstrated this approach on summarization, machine translation, arithmetic reasoning, and task-agnostic instruction tuning. The same principle has since been incorporated into the training pipelines of released LLM families: Gemma 2 performs teacher distillation on the student's distribution during supervised fine-tuning, while Qwen3 combines off-policy and on-policy strong-to-weak distillation in the post-training of its smaller models \citep{gemmateam2024gemma2,yang2025qwen3}.

On-Policy Self-Distillation (OPSD) is a recent instantiation of OPD that does
not require a distinct, typically larger, teacher model. Instead, the
trained student and the teacher are initialized from the same language
model. The student conditions only on the target problem and generates an
on-policy response, while the teacher additionally conditions on the target's
reference solution and provides next-token distributions along the student-generated prefixes. Across three Qwen3 model and several mathematics benchmarks, \targetopsd{} improves over the base model and supervised fine-tuning, while matching or exceeding GRPO with substantially
fewer sampled training tokens \citep{zhao2026opsd}. Conditioning the teacher on the solution may therefore produce more informative token-level targets for the student who does not see the solution. \textbf{This suggests an explanation for OPSD's gains: they arise because the teacher has privileged access to the answer and reasoning trace for the exact problem being solved by the student.}

This explanation is plausible, but the usual comparison does not highlight it. When a verified solution is placed in the teacher prompt, two things change simultaneously. First, it reveals target-specific privileged information: the
derivation and answer for the current problem. Second, it provides a complete worked solution within the teacher's context. This context may alter the teacher's next-token distribution due to its reasoning structure, notation, style, and effects on continuation and termination, even if it does not solve the current instance. Therefore, an improvement over the base model does not establish that OPSD transfers the privileged answer.

We ask whether OPSD depends on the identity of the solution, or whether part
of its effect arises from behavior induced by the teacher's additional
context. To probe this distinction, we introduce \method{}
(\emph{\textbf{O}n-\textbf{P}olicy \textbf{S}elf-\textbf{D}istillation from Other \textbf{P}roblems}). For a target
problem \(A\), standard OPSD gives the teacher the verified solution to
\(A\). \method{} instead gives it a different problem \(B\) and its
solution, while explicitly stating that the example is neither a solution
nor a hint for \(A\). The student still sees only \(A\). We otherwise
preserve the on-policy rollout, token-level objective,
and optimization procedure. This intervention breaks the target and reference
pairing while retaining the teacher-only worked-solution context
(Figure ~\ref{fig:teacher-context-swap}).
Across the evaluated Qwen3-1.7B, Qwen3-4B, and Qwen3-8B non-thinking models,
\method{} improves accuracy over the corresponding base model on
AIME 2024, AIME 2025, and HMMT 2025 (Section~\ref{sec:main_result}).
Therefore, the paired derivation and answer are not necessary for obtaining the OPSD-like improvements observed in these experiments.

Furthermore, we found that not all additional context yields a useful teacher. When evaluating the model on algebraic questions, we find that a worked solution drawn from the broader mathematical domain, rather than algebra specifically, provides a similar advantage. In contrast, replacing mathematical examples with physics problems and solutions significantly reduces accuracy and causes many generations to fall into repetitive, non-terminating trajectories. Thus, the identity and form of the teacher context matter, even though exact alignment with the target solution does not appear necessary.

Our contribution is therefore primarily diagnostic. We identify a confounding factor in the standard interpretation of OPSD, introduce a controlled intervention that decouples the target-reference pairing, and demonstrate that a significant amount of improvement persists despite this intervention. 
These findings suggest that OPSD
should not be interpreted solely as a privileged-answer transfer:
context-induced changes in teacher behavior may also constitute an important part of the signal being distilled.

\definecolor{promptred}{HTML}{B85F68}
\definecolor{promptgreen}{HTML}{4D806F}
\definecolor{promptink}{HTML}{182433}

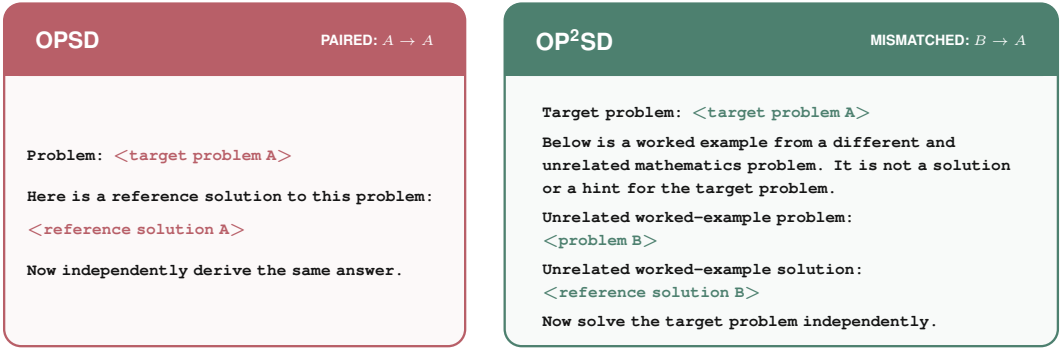
\begin{figure}[t]
\centering
\resizebox{\textwidth}{!}{%
\begin{tikzpicture}[x=1mm,y=1mm]
  \tikzset{
    outercard/.style={
      rounded corners=2.6mm,
      line width=1.0pt
    },
    prompttext/.style={
      anchor=north west,
      align=left,
      text=black!92,
      font=\ttfamily\bfseries\fontsize{7.0}{14}\selectfont
    }
  }

  \draw[
    outercard,
    draw=promptred,
    fill=promptred!4
  ] (-80,0) rectangle (-10,-52);
  \fill[
    promptred,
    rounded corners=2.3mm
  ] (-80,0) rectangle (-10,-11);
  \fill[promptred] (-80,-8.7) rectangle (-10,-11);

  \node[
    anchor=west,
    text=white,
    font=\sffamily\bfseries\fontsize{9.8}{11.2}\selectfont
  ] at (-76.5,-5.5) {OPSD};
  \node[
    anchor=east,
    text=white!86,
    font=\sffamily\bfseries\fontsize{6.1}{7.2}\selectfont
  ] at (-13.5,-5.5) {PAIRED: \(A\rightarrow A\)};

  \node[
    prompttext,
    anchor=north,
    text width=63mm
  ] at (-45,-21.0)
  {Problem: \textcolor{promptred}{\textless target problem A\textgreater}\\[1.3mm]
   Here is a reference solution to this problem:\\
   \textcolor{promptred}{\textless reference solution A\textgreater}\\[1.3mm]
   Now independently derive the same answer.};

  \draw[
    outercard,
    draw=promptgreen,
    fill=promptgreen!4
  ] (-4,0) rectangle (80,-52);
  \fill[
    promptgreen,
    rounded corners=2.3mm
  ] (-4,0) rectangle (80,-11);
  \fill[promptgreen] (-4,-8.7) rectangle (80,-11);

  \node[
    anchor=west,
    text=white,
    font=\sffamily\bfseries\fontsize{10.2}{11.6}\selectfont
  ] at (-0.5,-5.5) {OP\textsuperscript{2}SD};
  \node[
    anchor=east,
    text=white!86,
    font=\sffamily\bfseries\fontsize{6.1}{7.2}\selectfont
  ] at (76.5,-5.5) {MISMATCHED: \(B\rightarrow A\)};

  \node[
    prompttext,
    text width=76mm,
   font=\ttfamily\bfseries\fontsize{7.0}{10}\selectfont
  ] at (0.5,-14.5)
  {Target problem: \textcolor{promptgreen}{\textless target problem A\textgreater}\\[0.9mm]
   Below is a worked example from a different and\\
   unrelated mathematics problem. It is not a solution\\
   or a hint for the target problem.\\[0.9mm]
   Unrelated worked-example problem:\\
   \textcolor{promptgreen}{\textless problem B\textgreater}\\[0.9mm]
   Unrelated worked-example solution:\\
   \textcolor{promptgreen}{\textless reference solution B\textgreater}\\[0.9mm]
   Now solve the target problem independently.};

\end{tikzpicture}%
}
\caption{\textbf{Teacher prompt comparison.}
OPSD conditions the teacher on the reference paired
with target \(A\); \method{} substitutes a worked solution to a
different problem \(B\). The target, student rollout, teacher model, and
token-level objective are unchanged.}
\label{fig:prompt-comparison}
\end{figure}


\section{Problem Formulation}

\subsection{On-Policy Self-Distillation}

On-Policy Self-Distillation (OPSD) trains a language model using token-level
supervision from a frozen copy of its own initialization
\citep{zhao2026opsd}. The student and teacher, therefore, originate from the
same base model, but they receive different conditioning information. The
trainable student sees only the target problem and generates the trajectory
used for training. The teacher sees the same target problem together
with its reference solution and evaluates the student's trajectory under this
additional context. The method is \emph{on-policy} because supervision is
provided on prefixes generated by the current student, and it is
\emph{self-distillation} because the student and teacher are derived from the
same model rather than from models of different sizes.

Let $\mathcal D=\{(x_i,y_i^\star)\}_{i=1}^{N}$
denote a dataset of mathematical problems \(x_i\) paired with reference
solutions \(y_i^\star\). Let \(p_0\) be the base model and
\(p_\theta\) the trainable student initialized from \(p_0\). For target
problem \(x_i\), the student policy conditions only on the problem, whereas \targetopsd{} teacher additionally conditions on the paired
reference:
\begin{equation}
  p_S(\cdot\mid x_i)
  =p_\theta(\cdot\mid x_i), \qquad
  p_T^{\mathrm{target}}(\cdot\mid x_i,y_i^\star)
  =p_0(\cdot\mid x_i,y_i^\star).
  \label{eq:clean-policies}
\end{equation}
The teacher does not generate an alternative solution.
Thus, the teacher's advantage does not derive from having a larger parameter
count, but from its privileged access to \(y_i^\star\). This reference is
available only to the teacher during training; the student must solve the
problem without it both when generating training trajectories and at
inference time.
Given the two context-conditioned policies above, an \targetopsd{} training step
proceeds in two stages.

First, the current student samples a complete response
to the target problem:
\begin{equation}
  \hat y_i\sim p_\theta(\cdot\mid x_i).
  \label{eq:onpolicy-response}
\end{equation}

Second, to compute the OPSD loss, the same output is replayed under both
the student and the teacher. Note that the student prompt \(x_i\) and the teacher
prompt \((x_i,y_i^\star)\) are independently right-padded within their
respective minibatches before \(\hat y_i\) is appended; details are provided in
Appendix~\ref{sec:left-padding-control}.

\subsection{Target-Specific Privilege or Context-Induced Teacher Behavior?}

Existing accounts of OPSD, consistent with the broader literature on
learning using privileged information~\citep{VAPNIK2009544}, typically
attribute its gains to the teacher's access to privileged information about
the current problem~\citep{star2022,rest2023,mitra2025semantic,qi2025ice,
zhao2026opsd,hubotter2026reinforcement}. Under this interpretation, the
teacher receives the verified (partial) answer and reasoning trace for the same problem (also known as ``reference solution'' or ``target solution'')
that the student is attempting to solve, and the student learns from the
teacher's predictions under this target-specific information.

However, providing a reference solution changes not only what the teacher
knows about the target, but also the context under which it produces
token-level supervision. A worked solution may alter the teacher's
next-token distribution through its reasoning structure, notation, style,
and effects on continuation and termination, even when it conveys no
information about the current instance. We therefore distinguish
\emph{target-specific privilege} from \emph{context-induced teacher
behavior}. Our central question is whether OPSD requires privileged
information about the current target, or whether a teacher-only
worked-solution context can induce a useful supervisory distribution without
revealing the target's solution.

\section{\texorpdfstring{\method}{OP2SD}: On-Policy Self-Distillation from Other Problems}

\subsection{Replacing the Paired Reference}

\method{} retains the on-policy training procedure of \targetopsd{} but changes the
source of the teacher-only reference context. In OPSD, the
teacher receives the reference solution paired with the current target.
In \method, that paired reference is replaced by a problem and solution pair from
a different training instance. This intervention is designed to test whether
the alignment of the instance-level between the target problem and the privileged
reference is necessary for the \targetopsd{} gain.

Formally, for each target problem \(x_i\), we select another problem \(x_j\)
and its reference solution \(y_j^\star\), subject to \(j\neq i\).
We refer to the pair \((x_j,y_j^\star)\) as the \emph{worked-example}.
The student policy is unchanged; on the other hand, the teacher instead conditions on both the target problem and the
worked-example (Fig.~\ref{fig:prompt-comparison}):
\begin{equation}
  p_T^{\mathrm{other}}(\cdot\mid x_i,x_j,y_j^\star)
  =
  p_0(\cdot\mid x_i,x_j,y_j^\star).
  \label{eq:other-problem-policy}
\end{equation}
As in the target solution OPSD, the teacher does not generate a separate response.
The paired target solution \(y_i^\star\) is not included in the conditioning
context of either policy. The teacher's prompt explicitly identifies
\((x_j,y_j^\star)\) as a worked example from a different problem and states
that it should not be treated as a solution or hint for \(x_i\).

\subsection{Distillation Objective}

\targetopsd{} and \method{} use the same rollout construction and
token-level distillation objective. To express both conditions in a common
form, let
\[
  m\in\{\mathrm{target},\mathrm{other}\},
  \qquad
  c_i^{\mathrm{target}}=y_i^\star,
  \qquad
  c_i^{\mathrm{other}}=(x_j,y_j^\star).
\]
For either condition, the current student first samples a response, and the student and teacher then evaluate the same student-generated prefixes.
The corresponding objective is
\begin{equation}
  \mathcal L_m(\theta)
  =
  \mathbb E_{(x_i,y_i^\star)\sim\mathcal D}
  \mathbb E_{\hat y_i\sim p_S(\cdot\mid x_i)}
  \left[
    \frac{1}{|\hat y_i|}
    \sum_{t=1}^{|\hat y_i|}
    D\!\left(
      p_T^m(\cdot\mid x_i,c_i^m,\hat y_{i,<t})
      \,\middle\|\,
      p_S(\cdot\mid x_i,\hat y_{i,<t})
    \right)
  \right].
  \label{eq:shared-distillation-objective}
\end{equation}
Here, \(D\) denotes the divergence at the token-level between the next-token distributions of the teacher and student
. In our experiments, both conditions use the same
clipped Forward-KL surrogate. The teacher distribution is treated as a fixed
target and the gradients are propagated only through the student. 
Thus, the two conditions differ in the teacher-only context \(c_i^m\):
\targetopsd{} uses the reference paired with the target, whereas
\method{} uses the worked-example from another problem. The student input,
rollout construction, teacher model, distillation objective, and optimization
procedures are otherwise unchanged.

\section{Experiments}
\label{sec:experimental-setup}

We evaluate \method{} using Qwen3-1.7B ($p_S$: non-thinking; $p_T$: thinking), Qwen3-4B ($p_S$, $p_T$: non-thinking), and Qwen3-8B ($p_S$, $p_T$: non-thinking)~\citep{yang2025qwen3}.
 All main experiments are conducted on the OpenThoughts Math dataset (\texttt{siyanzhao/Openthoughts\_math\_30k\_opsd}), following the experimental setup of \citet{zhao2026opsd}.
 For each model setting, we compare \method{} against \targetopsd{} and base model (Base). 
In \method{}, each target problem is paired with a worked-example drawn from
the same training dataset. The worked-example pool consists of entries
labeled \texttt{amc\_aime} or \texttt{aops\_forum} that do not include AIME2024, AIME2025, and HMMT Feb 2025 in the OpenThoughts Math
dataset, and each selected example includes both a problem and its reference solution.
The details of the experiments are provided in
Appendix~\ref{sec:detailed-experimental-settings}.
We evaluated these methods on AIME 2024, AIME 2025, and HMMT Feb 2025 \citep{balunovic2025matharena}. For each problem, we generate 12
outputs under each of four different seeds.


\subsection{Main results}\label{sec:main_result}

We report Avg@12 as the mean over four different seeds, together with a
corrected Monte Carlo standard error. Pass@12 and Vote@12 are reported as
means together with empirical seed standard errors. The definitions of Avg@12, Pass@12, and Vote@12 are given in
Appendix~\ref{sec:evaluation-uncertainty}.

\begin{table}[t]
\centering
\caption{\textbf{Main benchmark results.}
Avg@12 is reported as the mean ($\pm$) corrected Monte Carlo standard error,
while Pass@12 and Vote@12 are reported as the mean ($\pm$) empirical
decoding-seed standard error. Within each model and benchmark, the highest
Avg@12 point estimate is shown in \textbf{bold}, and the second-highest is
\underline{underlined}. \method{} achieves the highest Avg@12 point estimate
in eight of the nine model--benchmark groups.}
\label{tab:main-results}
\small
\setlength{\tabcolsep}{5pt}
\begin{tabular}{@{}lllrrr@{}}
\toprule
Model & Benchmark & Method
& Avg@12 & Pass@12 & Vote@12 \\
\midrule

\multirow{9}{*}{Qwen3-1.7B}
& \multirow{3}{*}{AIME 2024}
& Base
& \(49.86\pm0.94\)
& \(76.67\pm0.00\)
& \(70.00\pm2.72\) \\
& & \targetopsd
& \(\mathbf{55.42\pm0.82}\)
& \(78.33\pm0.96\)
& \(65.00\pm1.67\) \\
& & \method
& \(\underline{55.35\pm0.83}\)
& \(75.83\pm0.83\)
& \(70.00\pm2.36\) \\
\cmidrule(lr){2-6}

& \multirow{3}{*}{AIME 2025}
& Base
& \(37.36\pm0.81\)
& \(70.00\pm3.04\)
& \(48.33\pm0.96\) \\
& & \targetopsd
& \(\underline{40.35\pm0.77}\)
& \(65.00\pm2.15\)
& \(50.00\pm1.36\) \\
& & \method
& \(\mathbf{40.69\pm0.74}\)
& \(63.33\pm2.36\)
& \(52.50\pm1.60\) \\
\cmidrule(lr){2-6}

& \multirow{3}{*}{HMMT 2025}
& Base
& \(23.61\pm0.64\)
& \(52.50\pm2.50\)
& \(28.33\pm0.96\) \\
& & \targetopsd
& \(\underline{25.76\pm0.63}\)
& \(50.83\pm2.10\)
& \(29.17\pm0.83\) \\
& & \method
& \(\mathbf{27.57\pm0.66}\)
& \(51.67\pm1.67\)
& \(30.83\pm0.83\) \\

\midrule

\multirow{9}{*}{\shortstack[l]{Qwen3-4B}}
& \multirow{3}{*}{AIME 2024}
& Base
& \(23.19\pm0.72\)
& \(50.83\pm1.60\)
& \(33.33\pm2.36\) \\
& & \targetopsd
& \(\underline{30.76\pm0.77}\)
& \(59.17\pm0.83\)
& \(45.83\pm1.60\) \\
& & \method
& \(\mathbf{31.53\pm0.80}\)
& \(60.83\pm1.60\)
& \(43.33\pm1.36\) \\
\cmidrule(lr){2-6}

& \multirow{3}{*}{AIME 2025}
& Base
& \(21.11\pm0.64\)
& \(46.67\pm2.72\)
& \(26.67\pm2.36\) \\
& & \targetopsd
& \(\underline{23.06\pm0.65}\)
& \(50.00\pm1.36\)
& \(28.33\pm0.96\) \\
& & \method
& \(\mathbf{30.62\pm0.71}\)
& \(55.00\pm2.15\)
& \(40.83\pm1.60\) \\
\cmidrule(lr){2-6}

& \multirow{3}{*}{HMMT 2025}
& Base
& \(11.67\pm0.49\)
& \(22.50\pm2.10\)
& \(16.67\pm0.00\) \\
& & \targetopsd
& \(\underline{15.42\pm0.57}\)
& \(35.83\pm2.85\)
& \(20.83\pm0.83\) \\
& & \method
& \(\mathbf{16.11\pm0.60}\)
& \(38.33\pm2.15\)
& \(20.00\pm1.92\) \\

\midrule

\multirow{9}{*}{\shortstack[l]{Qwen3-8B}}
& \multirow{3}{*}{AIME 2024}
& Base
& \(28.47\pm0.75\)
& \(58.33\pm3.19\)
& \(41.67\pm3.97\) \\
& & \targetopsd
& \(\underline{45.00\pm0.86}\)
& \(74.17\pm1.60\)
& \(61.67\pm2.15\) \\
& & \method
& \(\mathbf{55.83\pm0.87}\)
& \(80.00\pm1.36\)
& \(71.67\pm2.15\) \\
\cmidrule(lr){2-6}

& \multirow{3}{*}{AIME 2025}
& Base
& \(20.97\pm0.63\)
& \(44.17\pm2.10\)
& \(28.33\pm0.96\) \\
& & \targetopsd
& \(\underline{32.15\pm0.76}\)
& \(58.33\pm2.89\)
& \(40.00\pm1.36\) \\
& & \method
& \(\mathbf{42.85\pm0.71}\)
& \(62.50\pm2.10\)
& \(52.50\pm2.50\) \\
\cmidrule(lr){2-6}

& \multirow{3}{*}{HMMT 2025}
& Base
& \(11.81\pm0.51\)
& \(23.33\pm0.00\)
& \(16.67\pm0.00\) \\
& & \targetopsd
& \(\underline{16.18\pm0.60}\)
& \(39.17\pm1.60\)
& \(20.00\pm1.36\) \\
& & \method
& \(\mathbf{25.56\pm0.72}\)
& \(55.00\pm1.67\)
& \(30.83\pm2.10\) \\

\bottomrule
\end{tabular}
\end{table}

Table~\ref{tab:main-results} shows the main benchmark results. For
Qwen3-1.7B, \targetopsd{} and \method{} improve Avg@12 over Base on all three
benchmarks. These results suggest that access to paired target references is not necessary to achieve Avg@12 gains comparable to those of OPSD.
Avg@12 and Pass@12 do not always exhibit the same pattern. Although \method{} improves Avg@12 over Base on all three benchmarks for Qwen3-1.7B, its Pass@12 point estimates are slightly lower.
This pattern is consistent with \method{} increasing the probability of generating correct responses for problems that the base model can already solve occasionally, rather than uniformly expanding the set of problems solved within 12 samples (Appendix~\ref{sec:problem-level-agreement}).
The Qwen3-4B and Qwen3-8B results further demonstrate the effectiveness of \method{} at larger model scales, as it achieves the highest Avg@12 point estimate on all three benchmarks for both model sizes. For Qwen3-8B, it outperforms \targetopsd{} in Avg@12 by 10.83, 10.69, and 9.38 points on AIME 2024, AIME 2025, and HMMT 2025, respectively. We next examine whether these gains require additional generation.

\subsubsection{Accuracy under fixed token budgets}
\label{sec:accuracy-token-budget}

\begin{figure}[t]
  \centering
  \includegraphics[width=\textwidth]{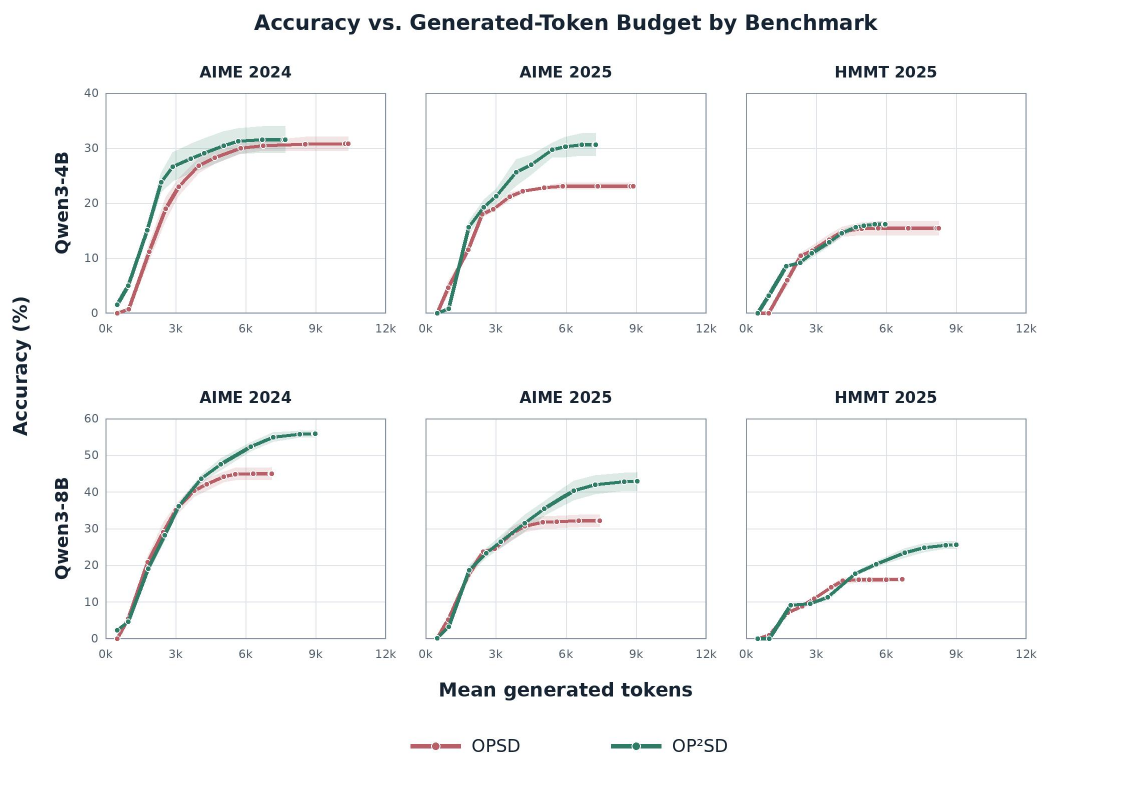}
  \caption{\textbf{Accuracy under generated-token budgets.}
  Each point is obtained by truncating the saved response at a fixed token
  cap and reapplying the original answer extractor and grader. If the model has not produced an answer by the truncation point, the response is counted as incorrect.
  The horizontal
  coordinate is the actual mean number of retained tokens after accounting
  for responses that terminate before the cap. }
  \label{fig:accuracy-token-budget}
\end{figure}

Using the saved Qwen3-4B and Qwen3-8B outputs, we truncate each response at
budgets from 500 to 38,912 tokens and reapply the original boxed answer
evaluator; prefixes without a complete boxed answer are counted as incorrect.
Figure~\ref{fig:accuracy-token-budget} shows
accuracy against the actual mean number of retained tokens, accounting for
responses that terminate before the specified budget.

Figure~\ref{fig:accuracy-token-budget} shows different scaling behavior. For
Qwen3-4B, \method{} remains above \targetopsd{} over nearly the entire budget
range and reaches a higher final accuracy with fewer mean generated tokens. Thus, the
4B gain is not explained by longer responses. For Qwen3-8B, the two methods
are comparable at small budgets, after which \method{} continues to improve
while \targetopsd{} saturates. Its final output is longer, but the additional
budget is accompanied by a substantially higher pooled accuracy. Hence,
\method{} does not uniformly shorten reasoning across model scales; rather, it achieves higher accuracy for a given token budget once the budget is sufficiently large.

The results reported so far show that paired target solutions are not necessary to obtain improvements comparable to those of OPSD. We next investigate which properties of the teacher signal account for this result. We first disentangle the effect of solution conditioning from that of the student-teacher thinking-mode asymmetry. We then consider a matched non-thinking setting and progressively examine whether the observed gains depend on diverse mathematical examples, broad domain alignment, a correct auxiliary derivation, or simply the presence of arbitrary additional context.

\paragraph{A mode-asymmetry inherited from the 1.7B OPSD setup.}

For Qwen3-1.7B, we adopted the original OPSD settings of \citet{zhao2026opsd}, in which the student generates in non-thinking mode while the frozen teacher scores the same trajectory in thinking mode.
The asymmetry is therefore inherited from the baseline recipe rather than introduced by \method{}, and retaining it enables a direct comparison with the original setting. 
It nevertheless confounds the role of worked-solution
context: an improvement could arise from the solution, the stronger teacher mode, generic on-policy distillation, or their interaction.

We introduce two controls to separate this
mode asymmetry from the effect of worked-solution context. In
\emph{Target-only}, the teacher receives the same problem-only content as
the student, with no reference solution. In \emph{Answer-only}, the teacher
additionally receives the final answer, but not its
derivation.
We confirm that both Target-only and Answer-only improve the base model's accuracy across all three benchmarks (see Appendix~\ref{sec:target-only-control}). Thus, in the 1.7B setting, improvements can arise even when the teacher receives neither a worked solution nor additional mathematical context.
Therefore, these gains cannot be exclusively attributed to the content of the privileged solution and may instead reflect, at least in part, the distillation of the behavior induced by the teacher’s thinking mode.
Complete results and prompt details are reported in
Appendix~\ref{sec:target-only-control}. To remove this asymmetry, we evaluate the controls using Qwen3-4B, with both the student and the teacher in non-thinking mode.
\begin{table}[t]
\centering
\caption{\textbf{Worked-solution controls with Qwen3-4B.}
Target-only provides no additional information beyond the target problem, whereas Answer-only additionally provides the verified final answer without a derivation. Unlike in the Qwen3-1.7B setting, neither control improves over Base: Target-only substantially degrades Avg@12 on all three benchmarks, and Answer-only partially recovers performance but remains below Base. These results indicate that the gains observed in the asymmetric 1.7B controls are not observed when student and teacher modes are matched.}
\label{tab:4b-solution-content-control}
\begin{tabular}{@{}lrrr@{}}
\toprule
Teacher information & AIME 2024 & AIME 2025 & HMMT 2025 \\
\midrule
Target-only
& \(1.18\pm0.28\)
& \(0.49\pm0.18\)
& \(0.76\pm0.22\) \\
Answer-only
& \(18.68\pm0.55\)
& \(17.85\pm0.55\)
& \(9.24\pm0.49\) \\
\midrule
Base
& \(23.19\pm0.72\)
& \(21.11\pm0.64\)
& \(11.67\pm0.49\) \\
\bottomrule
\end{tabular}
\end{table}

As shown in Table~\ref{tab:4b-solution-content-control}, the Qwen3-4B non-thinking setting behaves differently from the 1.7B setting.
Target-only performs substantially below Base on all three benchmarks.
Providing the verified final answer recovers part of this loss, but
Answer-only remains \(4.51\), \(3.26\), and \(2.43\) percentage points below
Base on AIME 2024, AIME 2025, and HMMT 2025, respectively. In contrast, both
target-solution \targetopsd{} and worked-problem \method{} improve over Base,
as shown in Table~\ref{tab:main-results}.

Thus, in this mode-matched configuration, 
improvement is observed only when the teacher is provided with a complete mathematical worked solution.
We therefore next investigate which properties of this
worked-solution context are responsible for the improvement.

\subsection{What Properties of the Other-Problem Context Matter?}
\label{sec:fixed-context-ablation}

\paragraph{Is diversity among worked examples necessary?}

The \method{} condition exposes the teacher to examples drawn from different problems. One possibility is that its benefit arises from the diversity of these examples. To test this hypothesis, we remove this diversity by conditioning the teacher on a single fixed algebra problem from Omni-MATH~\citep{gao2024omnimath} throughout training.
We refer to this condition as \emph{Fixed correct}. The exact problem and solution are provided in Appendix~\ref{sec:fixed-context-examples}. 
\begin{table}[t]
\centering
\caption{\textbf{Effect of diversity and form of the mathematical teacher context.}
Replacing the varying worked examples with a single fixed algebra problem and a concise, correct solution preserves and slightly improves Avg@12 across all three benchmarks with Qwen3-4B, showing that diversity among worked examples is unnecessary. A locally corrupted solution performs comparably to or better than the concise correct solution, whereas a substantially more verbose correct solution yields lower Avg@12 and shorter student responses. In contrast, a fixed trivial $1+1$ example performs poorly, indicating that an arbitrary correct mathematical example is not sufficient. Solution tokens denote the length of the teacher-provided solution, and Mean tokens denote the average student response length across all generations. The exact fixed contexts are provided in Appendix~\ref{sec:fixed-context-examples}.}

\label{tab:fixed-context-form-ablation}
\small
\setlength{\tabcolsep}{4.5pt}
\resizebox{\textwidth}{!}{%
\begin{tabular}{@{}lrrrrr@{}}
\toprule
\shortstack{Teacher context}
& \shortstack{Solution tokens}
& AIME 2024
& AIME 2025
& HMMT 2025
& \shortstack{Mean tokens} \\
\midrule
Varying mathematics examples
& --
& \(31.53\pm0.80\)
& \(30.62\pm0.71\)
& \(16.11\pm0.60\)
& 6,979 \\
\midrule
Fixed correct
& 204
& \(35.83\pm0.87\)
& \(31.94\pm0.79\)
& \(17.71\pm0.62\)
& 6,428 \\
Fixed locally corrupted
& 204
& \(36.04\pm0.83\)
& \(34.86\pm0.78\)
& \(17.99\pm0.63\)
& 6,442 \\
Fixed verbose correct
& 742
& \(33.89\pm0.85\)
& \(28.40\pm0.74\)
& \(15.76\pm0.68\)
& 5,773 \\
Fixed trivial correct (\(1+1\))
& 18
& \(5.21\pm0.52\)
& \(5.28\pm0.49\)
& \(2.29\pm0.35\)
& 19,029 \\
\bottomrule
\end{tabular}%
}
\end{table}
Table~\ref{tab:fixed-context-form-ablation} shows that the fixed-example run
attains Avg@12 point estimates of \(35.83\), \(31.94\), and \(17.71\) on
AIME 2024, AIME 2025, and HMMT 2025, compared with \(31.53\), \(30.62\),
and \(16.11\) for varying-example \method{} with Qwen3-4B. Thus, an OPSD-like improvement
can occur even when the auxiliary context is reduced to one repeated example.

We next hold the fixed worked example constant while varying its solution. The Fixed locally corrupted condition uses a solution with the same length as the concise correct version (204 tokens) but introduces a local error that changes the final answer. The Fixed verbose correct condition instead preserves the correct answer while deliberately making the derivation verbose by increasing it to 742 tokens.
The Fixed locally corrupted condition yields Avg@12 point estimates that are comparable to or higher than those of the Fixed correct condition on all three benchmarks. This result suggests that \method{} does not require the solution to be fully correct to achieve strong performance.
The verbose correct context exhibits a different pattern. Relative to the concise correct context, it lowers Avg@12 by $1.94$, $3.54$, and $1.95$ points while reducing the pooled mean student response length by approximately ($10.2\%$). In this experiment, these results suggest that a concise, focused solution example may provide a more effective teacher signal than a substantially expanded derivation. More broadly, the form and degree of compression of the teacher-only context appear to be associated with both accuracy and student response length, although these comparisons do not identify the underlying causal mechanism.

As an additional control, we condition the teacher on a fixed, trivially correct (1+1) example (see Appendix~\ref{sec:fixed-context-examples}). This condition performs poorly, achieving Avg@12 scores of only $5.21$, $5.28$, and $2.29$ on AIME 2024, AIME 2025, and HMMT 2025, respectively. Thus, although diversity is not necessary, merely providing an arbitrary correct mathematical example is not sufficient to obtain the observed gains.

\paragraph{Is broad domain matching necessary?}

The fixed-example result implies that \method{} does not necessarily require the teacher to observe a diverse stream of worked-problem solutions to attain strong performance. However, the successful fixed example is still mathematical and may benefit from belonging to the same broad domain as the target problems. We therefore next ask whether the auxiliary problem must match the target's mathematical domain.

Using Omni-MATH
\citep{gao2024omnimath}, we train on 1,280 Algebra problems and evaluate on
30 held-out Algebra problems. The teacher receives another Algebra
problem and its solution or a Geometry problem and its solution. The two other problem sets are matched by source,
difficulty, and context length; construction and leakage checks are provided
in Appendix~\ref{sec:domain-control-details}.
\begin{table}[t]
\centering
\caption{\textbf{Effect of coarse mathematical domain matching.}
Using Qwen3-4B in the matched non-thinking setting, we train on Algebra targets and compare teacher-only contexts containing either another Algebra problem and solution or a matched Geometry problem and solution. Geometry context performs comparably to, and slightly better than, Algebra context on the 30 held-out Algebra problems, providing no clear evidence that matching the target's coarse mathematical domain is necessary.  Dataset construction and leakage checks are provided in Appendix~\ref{sec:domain-control-details}.}
\label{tab:domain-control-main}
\setlength{\tabcolsep}{7pt}
\begin{tabular}{@{}lrr@{}}
\toprule
Teacher-only context & Avg@12 & Pass@12 \\
\midrule
Another Algebra problem and solution
& \(57.43\pm0.71\) & \(79.17\pm0.83\) \\
Another Geometry problem and solution
& \(59.10\pm0.73\) & \(80.00\pm1.36\) \\
\bottomrule
\end{tabular}
\end{table}
From Table~\ref{tab:domain-control-main}, the Geometry condition is \(1.67\) points higher, but this apparent advantage
is driven by one target near the boundary between Algebra and analytic
Geometry. Excluding that problem reverses the ordering: Algebra obtains
\(59.41\%\) and Geometry obtains \(58.98\%\) Avg@12. We therefore find no
clear evidence that matching a coarse domain label is necessary.

\paragraph{Is mathematical context necessary?}

The preceding controls relax target alignment, example diversity, broad
domain matching, and, in one pilot, exact correctness without uniformly
eliminating the gain. This does not imply that any additional text produces a
useful teacher. For this cross-subject control, we sample 2,768
problem and solution pairs from the external CAMEL Physics dataset
\citep{li2023camel}, matching the size of the mathematical example pool. We
keep the OpenThoughts Math target stream unchanged and replace only the
teacher-only worked-example context. Data construction and the complete
teacher-prompt template are reported in
Appendix~\ref{sec:physics-donor-control}.

\begin{table}[t]
\centering

\caption{\textbf{Cross-subject control for mathematical solution context.}
Using Qwen3-4B in the matched non-thinking setting, we replace only the teacher's mathematical worked examples with CAMEL Physics problem and solution pairs. Mathematical contexts improve Avg@12 over Base on all three benchmarks, whereas physics contexts fall below Base throughout. Thus, an arbitrary cross-subject context does not preserve the \method{} gain, suggesting that the mathematical solution context, or an associated property, is important. Additional metrics and implementation details are provided in Appendix~\ref{sec:physics-donor-control}.}
\label{tab:physics-control-main}
\setlength{\tabcolsep}{7pt}
\begin{tabular}{@{}lrrr@{}}
\toprule
Condition & AIME 2024 & AIME 2025 & HMMT 2025 \\
\midrule
Base
& \(23.19\pm0.72\) & \(21.11\pm0.64\) & \(11.67\pm0.49\) \\
\method{} (mathematics)
& \(31.53\pm0.80\) & \(30.62\pm0.71\) & \(16.11\pm0.60\) \\
\method{} (physics)
& \(18.68\pm0.69\) & \(20.21\pm0.61\) & \(8.68\pm0.50\) \\
\bottomrule
\end{tabular}
\end{table}

As shown in Table~\ref{tab:physics-control-main}, replacing mathematical
worked solutions with physics examples reduces Avg@12 from \(31.53\) to
\(18.68\) on AIME 2024, from \(30.62\) to \(20.21\) on AIME 2025, and from
\(16.11\) to \(8.68\) on HMMT 2025. The physics condition is also below Base
on all three benchmarks. Appendix~\ref{sec:physics-donor-control} reports the
corresponding Pass@12, Vote@12, and answer-format validity, together with the
actual teacher-prompt template; these diagnostics show that the accuracy drop
is accompanied by substantially less reliable answer formatting. 
Together, these results suggest that the mathematical solution context, or some property associated with it, is important for obtaining the \method{} gain. However, because the physics condition differs from the mathematics condition in several respects, this experiment does not establish that mathematical subject matter itself is the causal factor.

Overall, strong point estimates persist after removing instance-level alignment, diversity in worked examples, and coarse domain matching. However, the gain from \method{} is substantially reduced when mathematical worked solutions are replaced with worked solutions from another subject. 
These experiments do not identify a unique mechanism, but they shift the explanation away from direct privileged-answer transfer and toward teacher behavior elicited by a mathematical worked-solution context. Taken together, the results indicate that exposure to an example of a substantive mathematical solution, rather than an arbitrary additional context, is important to obtain the gain of \method{}.

\section{Related Work}

\paragraph{On-policy distillation.}
Knowledge distillation conventionally trains a student to match a teacher's
predictive distribution \citep{hinton2015distillation}. Sequence-level
distillation instead trains an autoregressive student on sequences decoded by
the teacher \citep{kim2016sequence}, but fixed teacher trajectories do not
cover all prefixes that the student encounters at inference time. MiniLLM
addresses this mismatch by optimizing a reverse-KL objective on student
samples \citep{gu2024minillm}. Generalized Knowledge Distillation (GKD) likewise queries the teacher on student-generated sequences and supports different divergences and mixtures of on-policy student outputs and fixed off-policy sequences \citep{agarwal2024onpolicy}. DistiLLM combines a skew-KL objective with an
adaptive off-policy scheme for using student-generated outputs more
efficiently \citep{ko2024distillm}.

\paragraph{Reasoning supervision and privileged information.}
One way to transfer reasoning is to use teacher rationales as explicit
supervised targets, as in Distilling Step-by-Step
\citep{hsieh2023distilling}. \targetopsd{} instead evaluates the student's own
trajectory with a token-level teacher distribution. Its asymmetric context is
closely related to learning using privileged information, where additional
training-time features are unavailable at inference
\citep{VAPNIK2009544,lopezpaz2016generalized}. In OPSD, a copy of the
model receives the verified solution to the current problem while the student
receives only the problem \citep{zhao2026opsd}. Related work on privileged-information distillation studies action-only transfer in multi-turn agentic environments, proposing both a jointly trained privileged teacher--student objective and a reverse-KL-regularized on-policy RL alternative, which the authors also call OPSD \citep{penaloza2026privileged}. These
approaches retain privileged information that is relevant to the current
instance or trajectory.

\paragraph{Context-conditioned self-teachers.}
Context distillation trains a model to retain behavior induced by
instructions, examples, or scratchpads after that context is removed
\citep{snell2022context}. Self-Distillation Fine-Tuning (SDFT) brings this
idea on-policy, using a demonstration-conditioned copy of the model to provide
supervision for continual skill and knowledge acquisition
\citep{shenfeld2026sdft}. On-Policy Context Distillation (OPCD) similarly
distills a context-conditioned teacher, but constructs its context from
extracted experience or optimized instructions \citep{ye2026opcd}. OPCD
reports that directly substituting raw
previous-problem traces for extracted experience can reduce math validation
accuracy. Self-Supervised On-Policy Distillation (SSOPD), by contrast, needs
no external solution trace: within a mixed group of rollouts, it conditions
the teacher on a short correct completion and distills that distribution into
prefixes of a persistent wrong completion \citep{tan2026self}. 

\paragraph{Recent analyses and variants of \targetopsd{}.}
Several recent variants reconsider which parts of the privileged teacher signal should be distilled. Purified OPSD decomposes the teacher distribution into a reference-induced component and a question-conditioned component intended to remain useful at inference time, and suppresses the former using an additional reference-only teacher \citep{shen2026purified}. ROSD argues that globally imitating a reference-conditioned teacher may encourage trajectory imitation rather than correction of the student's specific error, and instead applies distillation locally around an identified error \citep{zhao2026rosd}. PW-OPSD similarly questions uniform token-level supervision and assigns position-dependent weights according to the estimated reliability of the privileged teacher \citep{liu2026teacher}. These methods modify or filter the signal obtained from the target-aligned reference solution. 
\citet{kaur2026rethinking} report that privileged-context on-policy distillation can degrade the long-budget reasoning performance of thinking models, particularly when the teacher receives a full reference solution. These methods modify or filter supervision from a target-aligned reference.
In contrast, \method{} retains the same distillation rule and intervenes on
whether the worked solution is paired with the target.

\section{Limitations}

Our study is intentionally narrow. All target-side training and evaluation
tasks are mathematical, and the experiments use only Qwen3-1.7B, Qwen3-4B, and Qwen3-8B. 
Each trained condition is represented by one training run. The four seeds
measure sampling variation for a fixed trained model but do not capture
training-run variation. This limitation is particularly important for the
fixed, corrupted, and verbose contexts, which are obtained from separate
training runs.
Finally, we observe deterioration under longer \method{} training,
showing that a paired target solution is not required for this failure mode,
but we do not determine why longer OPSD-family training degrades. Establishing
that mechanism will require repeated training runs and controlled token-level
interventions.

\section{Conclusion}

We investigated whether OPSD requires the teacher to observe a verified solution to the same problem as the student. Across three model settings and three mathematics benchmarks, \method{} remains competitive with \targetopsd{}, showing that paired target solutions are not necessary for OPSD-like gains.
Most of the OPSD gain is retained, or even reinforced, after the reference solution to the target problem is replaced by the solution to a different problem (\method{}). Moreover, the gain persists when the diversity of the problem-solution pair given to the teacher is reduced to a single fixed example.
This shows that neither diverse auxiliary information nor exact contextual information is required. 
At the same time, trivial mathematical and cross-subject physics
contexts fail to preserve the improvement, indicating that arbitrary
additional text is insufficient. The verbose contexts also reduced the effectiveness of the teacher.
Taken together, these results imply that
the major source of the observed gain is not the privileged answer to the target problem, but the change in the teacher's token-level behavior induced by a mathematical context that works.

\ificlrfinal 
\section*{Acknowledgments}
J. Komiyama was supported by the MBZUAI Start-up Fund [BF0121].
\fi

\bibliographystyle{iclr2026_conference}
\bibliography{references}
\clearpage
\appendix

\section{Evaluation Metrics and Uncertainty}
\label{sec:evaluation-uncertainty}

Let \(M\) be the number of evaluation problems, \(S\) the number of decoding
seeds, and \(N\) the number of generations per problem and seed. In our main
experiments, \(M=30\), \(S=4\), \(N=12\), and hence each problem has
\(R=SN=48\) pooled generations. 

We report Avg@12 together with its corrected Monte Carlo standard error, computed as follows:
\begin{equation}
  \text{Monte Carlo standard error}  =
  \frac{100}{M}
  \sqrt{
    \frac{1}{R-1}
    \sum_{i=1}^{M}
    \hat p_i(1-\hat p_i)
  },
\end{equation}
where $M$ is the number of problems, $R$ is the number of generations per problem, and $\hat p_i$ is the fraction of correct generations for problem $i$. In particular, $M=30, R=4 \times 12$ in our experiments.

Pass@12 is computed separately for each seed as the percentage of problems
with at least one correct generation among that seed's \(N=12\) outputs.
Vote@12 is likewise computed per seed as the percentage of problems for
which the most frequent extracted, formatted answer among the 12 outputs is
graded correct. Let \(Z_s\) denote either the seed-level percentage and let
\(\bar Z=S^{-1}\sum_{s=1}^{S}Z_s\). We report \(\bar Z\) together with the
empirical decoding-seed standard error

\begin{equation}
  \widehat{\operatorname{SE}}_{\mathrm{seed}}(\bar Z)
  = \sqrt{
    \frac{1}{S(S-1)}
    \sum_{s=1}^{S}(Z_s-\bar Z)^2
  },
  \label{eq:seed-se}
\end{equation}

which is the sample standard deviation across decoding seeds divided by
\(\sqrt{S}\). These uncertainty estimates describe stochastic decoding for
a fixed trained model and fixed benchmark. They do not include variation
across training runs or uncertainty from sampling a different problem set.

\section{Detailed Experimental Settings}
\label{sec:detailed-experimental-settings}

All experiments are conducted with H$100$ $\times 4$ GPUs.


Tables~\ref{tab:training-setup}--\ref{tab:training-setup-8b} summarize the
model-specific training configurations.
\begin{table}[htbp]
\centering
\caption{Qwen3-1.7B training configuration shared by \targetopsd{} and
\method.}
\label{tab:training-setup}
\small
\begin{tabularx}{\textwidth}{@{}lX@{}}
\toprule
Component & Setting \\
\midrule
Base model & Qwen3-1.7B \\
Training dataset & \texttt{siyanzhao/Openthoughts\_math\_30k\_opsd} \\
Training seed & 42 \\
Per-device batch / gradient accumulation & 4 / 2 \\
Student and teacher mode & Non-thinking / thinking \\
Maximum student completion & 1,024 tokens \\
Student sampling & temperature 1.1, top-\(p\) 0.95, top-\(k\) 20 \\
Learning rate & \(5\times10^{-6}\), linear decay \\
Maximum gradient norm & 0.1 \\
Adapter & LoRA \(r=64\), \(\alpha=128\) \\
Teacher & Frozen base model; LoRA disabled \\
Objective & Forward KL, temperature 1.1, \(\beta=0\) \\
KL clip & \(\tau=0.05\) \\
\bottomrule
\end{tabularx}
\end{table}

\begin{table}[htbp]
\centering
\caption{Qwen3-4B non-thinking training configuration shared by
\targetopsd{} and \method{}.}
\label{tab:training-setup-4b}
\small
\begin{tabularx}{\textwidth}{@{}lX@{}}
\toprule
Component & Setting \\
\midrule
Base model & Qwen3-4B \\
Training dataset & \texttt{siyanzhao/Openthoughts\_math\_30k\_opsd} \\
Training seed & 42 \\
Per-device batch / gradient accumulation & 4 / 2 \\
Student and teacher mode & Non-thinking / non-thinking \\
Maximum student completion & 1,024 tokens \\
Student sampling & temperature 1.1, top-\(p\) 0.95, top-\(k\) 20 \\
Learning rate & \(5\times10^{-6}\), constant \\
Maximum gradient norm & 0.1 \\
Adapter & LoRA \(r=64\), \(\alpha=128\) \\
Teacher & Frozen base model; LoRA disabled \\
Objective & Forward KL, temperature 1.1, \(\beta=0\) \\
KL clip & \(\tau=10^{-6}\) \\
\bottomrule
\end{tabularx}
\end{table}

\begin{table}[htbp]
\centering
\caption{Qwen3-8B non-thinking training configuration shared by
\targetopsd{} and \method{}.}
\label{tab:training-setup-8b}
\small
\begin{tabularx}{\textwidth}{@{}lX@{}}
\toprule
Component & Setting \\
\midrule
Base model & Qwen3-8B \\
Training dataset & \texttt{siyanzhao/Openthoughts\_math\_30k\_opsd} \\
Training seed & 42 \\
Per-device batch / gradient accumulation & 2 / 4 \\
Student and teacher mode & Non-thinking / non-thinking \\
Maximum student completion & 1,024 tokens \\
Student sampling & temperature 1.1, top-\(p\) 0.95, top-\(k\) 20 \\
Learning rate & \(5\times10^{-6}\), constant \\
Maximum gradient norm & 0.1 \\
Adapter & LoRA \(r=64\), \(\alpha=128\) \\
Teacher & Frozen base model; LoRA disabled \\
Objective & Forward KL, temperature 1.1, \(\beta=0\) \\
 KL clip & \(\tau=10^{-7}\) \\
\bottomrule
\end{tabularx}
\end{table}

\clearpage

\section{Details of the Padding}
\label{sec:left-padding-control}

The main experiments follow the implementation of OPSD \citep{zhao2026opsd}. Student prompts
are right-padded to the longest student prompt in each minibatch, while teacher
prompts are padded independently to the longest teacher prompt. The sampled
response is then appended after the padded prompt. Consequently, shorter
examples contain masked padding between the prompt and response, and the
amount of padding can differ between the student and teacher inputs. Although
these padding tokens are masked out of attention, they affect the position
from which the response is scored. 
We therefore confirm the matched non-thinking Qwen3-4B comparison using
left-padded student and teacher prompts with the same checkpoint as Table~\ref{tab:main-results}. This shifts the padding to the start of each sequence, restoring adjacency between the unpadded prompt and the sampled response for both policies. This intervention removes padding from the boundary between the prompt and response.

\begin{table}[htbp]
\centering
\caption{\textbf{Qwen3-4B non-thinking results with left-padded loss
replay.} \method{} retains a substantial
improvement after right padding is removed.}
\label{tab:left-padding-control}
\begin{tabular}{@{}llrrr@{}}
\toprule
Benchmark & Method & Avg@12 & Pass@12 & Vote@12 \\
\midrule
\multirow{3}{*}{AIME 2024}
& Base          & \(23.19\pm0.72\) & \(50.83\pm1.60\) & \(33.33\pm2.36\) \\
& \targetopsd{} & \(27.78\pm0.88\) & \(64.17\pm1.60\) & \(52.50\pm0.83\) \\
& \method{}     & \(41.11\pm0.89\) & \(73.33\pm1.36\) & \(57.50\pm0.83\) \\
\midrule
\multirow{3}{*}{AIME 2025}
& Base          & \(21.11\pm0.64\) & \(46.67\pm2.72\) & \(26.67\pm2.36\) \\
& \targetopsd{} & \(19.44\pm0.53\) & \(36.67\pm2.36\) & \(23.33\pm2.36\) \\
& \method{}     & \(33.89\pm0.71\) & \(55.83\pm2.85\) & \(41.67\pm0.96\) \\
\midrule
\multirow{3}{*}{HMMT 2025}
& Base          & \(11.67\pm0.49\) & \(22.50\pm2.10\) & \(16.67\pm0.00\) \\
& \targetopsd{} & \(12.85\pm0.49\) & \(29.17\pm2.50\) & \(16.67\pm0.00\) \\
& \method{}     & \(17.36\pm0.64\) & \(45.00\pm3.97\) & \(22.50\pm1.60\) \\
\bottomrule
\end{tabular}
\end{table}

As shown in Table~\ref{tab:left-padding-control}, \method{} continues to yield substantial improvements under left padding, whereas the gains from \targetopsd{} are reduced, especially in AIME2025. 
Therefore, this setting also supports our main claim that access to the exact reference solution paired with the target is unnecessary.

\section{Detail of Result of Table~\ref{tab:domain-control-main}}
\label{sec:domain-control-details}

\junpei{It might be good to define the term for the problem in the main paper (donor is not defined in the main paper). Maybe "worked-example problem"?}
The worked-problem domain control uses a deterministic split of Omni-MATH
\cite{gao2024omnimath}, which does not provide an official train/test split
for this experiment. We select 1,280 Algebra-tagged, non-Geometry training
targets and 30 disjoint held-out targets, balanced across lower, middle, and
upper annotated difficulty. Each \method{} arm uses a 320-example
worked-example pool: Algebra/non-Geometry for \method-ID and Geometry-only
for \method-OOD. worked-examples share source provenance and are aligned by
difficulty and Qwen-tokenized context length; both use the same
deterministic target-to-context-example index mapping.
The 1.7B use training seed 42, 100 updates, an effective
batch of 32, and the remaining optimization settings in
Table~\ref{tab:training-setup}. This gives 3,200 target exposures, or 2.5
passes over the target set. We evaluate checkpoint 100 with 12 samples per
problem and 4 seeds.

We repeat the control with Qwen3-4B while disabling thinking for the student,
teacher, and evaluation. Table~\ref{tab:domain-control} reports the complete
results for both model settings.

\begin{table}[htbp]
\centering
\caption{\textbf{Complete coarse domain-matching control on a held-out Omni-MATH Algebra split.}
We evaluate Qwen3-1.7B and Qwen3-4B on 30 held-out Algebra problems, comparing teacher-only contexts drawn from either Algebra (\method-ID) or Geometry (\method-OOD). The worked-example pools are matched by source, difficulty, and context length. For Qwen3-1.7B, \method-ID exceeds \method-OOD by only (0.63) Avg@12 points, whereas for Qwen3-4B, \method-OOD is (1.67) points higher. The latter difference is driven by a single problem near the boundary between Algebra and analytic Geometry; excluding it reverses the ordering. Overall, the results provide no consistent evidence that matching the teacher-only context to the target's coarse mathematical domain is necessary.}
\label{tab:domain-control}
\small
\setlength{\tabcolsep}{4.5pt}
\begin{tabularx}{\textwidth}{@{}lXrrr@{}}
\toprule
Method & Teacher-only context & Avg@12 & Pass@12 & Vote@12 \\
\midrule
\multicolumn{5}{@{}l}{\textbf{Qwen3-1.7B}} \\
Base
& None
& \(64.72\pm0.65\) & \(77.50\pm1.60\) & \(70.00\pm2.36\) \\
\targetopsd{}
& Current target solution
& \(66.74\pm0.56\) & \(76.67\pm0.00\) & \(70.00\pm1.36\) \\
\method-ID
& Another Algebra problem and solution
& \(68.06\pm0.59\) & \(80.83\pm0.83\) & \(70.00\pm1.36\) \\
\method-OOD
& Another Geometry problem and solution
& \(67.43\pm0.56\) & \(76.67\pm0.00\) & \(70.83\pm0.83\) \\
\midrule
\multicolumn{5}{@{}l}{\textbf{Qwen3-4B}} \\
Base
& None
& \(56.60\pm0.76\) & \(79.17\pm2.10\) & \(65.83\pm2.50\) \\
\targetopsd{}
& Current target solution
& \(55.69\pm0.65\) & \(70.83\pm1.60\) & \(60.83\pm1.60\) \\
\method-ID
& Another Algebra problem and solution
& \(57.43\pm0.71\) & \(79.17\pm0.83\) & \(64.17\pm1.60\) \\
\method-OOD
& Another Geometry problem and solution
& \(59.10\pm0.73\) & \(80.00\pm1.36\) & \(62.50\pm0.83\) \\
\bottomrule
\end{tabularx}
\end{table}

With Qwen3-1.7B, \method-ID exceeds \method-OOD by only \(0.63\) points
in Avg@12 and wins, ties, and loses on 6, 19, and 5 targets, respectively.
The ordering reverses with Qwen3-4B, where OOD is \(1.67\) points higher.
This apparent OOD advantage is concentrated in one target asking about the
intersections of a circle and a parabola: OOD solves it in 30 of 48 samples,
whereas ID solves it in none. Although labeled Algebra, the problem lies near
the boundary with analytic Geometry. After excluding it, ID obtains
\(59.41\%\) Avg@12 and OOD obtains \(58.98\%\). We therefore find no
consistent evidence that matching these coarse domain labels is necessary.

\junpei{Updated the section title}
\section{Target-Only and Answer-Only Teacher Controls}
\label{sec:target-only-control}

Following \cite{zhao2026opsd}, in the Qwen3-1.7B, the student generates in non-thinking mode
while the teacher scores the same trajectory in thinking mode. 
This asymmetry suggests an alternative explanation for the gain: the student may learn from the behavior of a stronger inference mode even when the teacher has no access to privileged solution information.
We test this explanation with two controls. 
Namely, in the Target-only setting, the teacher receives only the target problem,
whereas in the Answer-only setting, the teacher additionally receives the verified final answer
$y_i^\star$ but withholds its derivation. The student prompt remains unchanged in each condition.

\begin{table}[t]
\centering
\caption{\textbf{Target-only and Answer-only controls across student--teacher mode settings.}
For Qwen3-1.7B, the student operates in non-thinking mode while the teacher evaluates the same trajectory in thinking mode. Under this asymmetric setting, both Target-only and Answer-only improve Avg@12 over Base on all three benchmarks, and revealing the verified final answer provides no consistent advantage over Target-only. For Qwen3-4B, both student and teacher use non-thinking mode; in this matched setting, neither control improves over Base, although Answer-only substantially outperforms Target-only. These results indicate that the gains of the 1.7B controls can arise without a worked-solution context and may depend on the student and teacher mode asymmetry.}
\label{tab:target-only-control}
\small
\setlength{\tabcolsep}{4.5pt}
\resizebox{\textwidth}{!}{%
\begin{tabular}{@{}llrrrrr@{}}
\toprule
Model & Benchmark & Base & Target-only & Answer-only & \targetopsd{} & \method{} \\
\midrule
\multirow{3}{*}{Qwen3-1.7B}
& AIME 2024
& \(49.86\pm0.94\)
& \(54.51\pm0.86\)
& \(53.82\pm0.88\)
& \(\mathbf{55.42\pm0.82}\)
& \(55.35\pm0.83\) \\
& AIME 2025
& \(37.36\pm0.81\)
& \(\mathbf{41.81\pm0.76}\)
& \(40.07\pm0.77\)
& \(40.35\pm0.77\)
& \(40.69\pm0.74\) \\
& HMMT 2025
& \(23.61\pm0.64\)
& \(26.67\pm0.64\)
& \(24.38\pm0.63\)
& \(25.76\pm0.63\)
& \(\mathbf{27.57\pm0.66}\) \\
\midrule
\multirow{3}{*}{\shortstack[l]{Qwen3-4B}}
& AIME 2024
& \(23.19\pm0.72\)
& \(1.18\pm0.28\)
& \(18.68\pm0.55\)
& \(30.76\pm0.77\)
& \(\mathbf{31.53\pm0.80}\) \\
& AIME 2025
& \(21.11\pm0.64\)
& \(0.49\pm0.18\)
& \(17.85\pm0.55\)
& \(23.06\pm0.65\)
& \(\mathbf{30.62\pm0.71}\) \\
& HMMT 2025
& \(11.67\pm0.49\)
& \(0.76\pm0.22\)
& \(9.24\pm0.49\)
& \(15.42\pm0.57\)
& \(\mathbf{16.11\pm0.60}\) \\
\bottomrule
\end{tabular}
}
\end{table}

Target-only remains above Base on all three Qwen3-1.7B. Adding
the verified answer does not improve over Target-only; its Avg@12 point
estimate is lower by \(0.69\), \(1.74\), and \(2.29\) percentage points on
AIME 2024, AIME 2025, and HMMT 2025, respectively. Thus, a substantial part
of the 1.7B improvement does not require a worked solution and is consistent
with generic on-policy distillation or transfer from the teacher's thinking
mode. This control does not establish that thinking capability itself is
transferred: the resulting models are evaluated with thinking enabled,
and the comparison does not isolate the mode change from the generic
self-distillation.

In contrast, in Qwen3-4B, in which both student and teacher are non-thinking, the results show a very different pattern: Target-only substantially underperforms Base, and Answer-only also falls below Base on all three benchmarks, while both \targetopsd{} and \method{} outperform Base.
In this configuration,
successful distillation is therefore associated with a complete mathematical
worked-solution context, but not with exact target alignment: \method{}
provides a solution to a different problem and still improves.
We regard this contrast as descriptive evidence only, since it does not isolate the factor responsible for the difference across model settings.

\junpei{updated title}
\section{Effect of Extended Training}
\label{sec:long-horizon-sensitivity}

We further train \method{} for 400 updates and evaluate its final
checkpoint on AIME 2024. Table~\ref{tab:long-horizon-sensitivity} reports the
four-seed averages for Base and the 100- and 400-update \method{} conditions.
The 400-update
checkpoint obtains \(51.39\) Avg@12, a decrease of \(3.96\) percentage points
from the 100-update result. It nevertheless remains \(1.53\) points above
Base. The change is not a uniform collapse: Pass@12 increases from \(75.83\)
to \(79.17\), and answer-format validity remains above \(99\%\), whereas
Vote@12 falls from \(70.00\) to \(64.17\). The observed deterioration is
therefore specific to sampled accuracy and voting consensus in this
evaluation.
\citet{kaur2026rethinking} reported that standard on-policy distillation can degrade with prolonged training, particularly when the teacher receives a complete reference solution.
Although \method{} provides the teacher with a complete mathematical solution, that solution corresponds to a different problem rather than the target itself.
Similar to \targetopsd{}, \method{}'s performance declines with extended training.

\begin{table}[htbp]
\centering
\caption{\textbf{Long-horizon checkpoint sensitivity of \method{} on AIME 2024 with Qwen3-1.7B.}
Extending training from 100 to 400 updates reduces Avg@12 from $55.35$ to $51.39$ and Vote@12 from $70.00$ to $64.17$, although the 400-update checkpoint remains above Base in Avg@12. In contrast, Pass@12 and answer-format validity increase at the later checkpoint, indicating that the decline is not uniform across metrics. }
\label{tab:long-horizon-sensitivity}
\small
\setlength{\tabcolsep}{4.5pt}
\begin{tabularx}{\textwidth}{@{}Xrrrr@{}}
\toprule
Condition & Avg@12 & Pass@12 & Vote@12 & Format \\
\midrule
Base
  & \(49.86\pm0.94\) & 76.67 & 70.00 & 98.68 \\
\method{}, checkpoint 100
  & \(55.35\pm0.83\) & 75.83 & 70.00 & 99.03 \\
\method{}, checkpoint 400
  & \(51.39\pm0.90\) & 79.17 & 64.17 & 99.51 \\
\bottomrule
\end{tabularx}
\end{table}

\junpei{Updated title and text}
\section{Problem-Level Analysis of Accuracy Gains}
\label{sec:problem-level-agreement}

In this section, we analyze the performance gains of \targetopsd{} and \method{} over Base on a problem-by-problem basis.
Aggregate Avg@12 can improve either (a) because a method solves problems that Base never solves or (b) because it increases the probability of producing a correct answer on problems that Base already solves occasionally.
We examine these possibilities for Qwen3-1.7B by pooling results across the four seeds.
For each problem \(i\) and
method \(m\), let \(c_i^{(m)}\in\{0,\ldots,48\}\) denote the number of correct
responses among 12 samples under each of four seeds.
Table~\ref{tab:problem-level-agreement} reports the Pearson correlation
between these problem-level correct counts for every pair of methods. It also
reports how often the first method has a higher, equal, or lower count than the second method across all 90 benchmark problems.
For methods \(a\) and \(b\) over \(N\) problems, we define
\[
r(a,b)=
\frac{\sum_{i=1}^{N}(c_i^{(a)}-\bar c^{(a)})(c_i^{(b)}-\bar c^{(b)})}
{\sqrt{\sum_{i=1}^{N}(c_i^{(a)}-\bar c^{(a)})^2}
 \sqrt{\sum_{i=1}^{N}(c_i^{(b)}-\bar c^{(b)})^2}},
\]
where \(\bar c^{(m)}\) is the mean correct count of method \(m\) over those
problems.
\begin{table}[htbp]
\centering
\caption{\textbf{Problem-level agreement among Qwen3-1.7B methods.}
Pearson correlations r are computed from correct counts out of 48 responses per
problem. ``All'' pools the 90 problems from the three benchmarks. W/T/L
compares the first-named method with the second across those 90 problems.}
\label{tab:problem-level-agreement}
\footnotesize
\setlength{\tabcolsep}{3.5pt}
\begin{tabularx}{\textwidth}{@{}Xrrrrc@{}}
\toprule
Pair (first vs.\ second) & AIME24 \(r\) & AIME25 \(r\) & HMMT25 \(r\) &
All \(r\) & W/T/L \\
\midrule
\targetopsd{} vs.\ Base
  & 0.942 & 0.974 & 0.974 & 0.966 & 39/30/21 \\
\method{} vs.\ Base
  & 0.953 & 0.976 & 0.970 & 0.969 & 40/31/19 \\
Target-only vs.\ Base
  & 0.967 & 0.964 & 0.983 & 0.973 & 39/33/18 \\
\method{} vs.\ \targetopsd{}
  & 0.988 & 0.987 & 0.992 & 0.990 & 31/39/20 \\
Target-only vs.\ \targetopsd{}
  & 0.970 & 0.967 & 0.978 & 0.973 & 26/42/22 \\
Target-only vs.\ \method{}
  & 0.989 & 0.989 & 0.988 & 0.989 & 25/36/29 \\
\bottomrule
\end{tabularx}
\end{table}

The methods largely agree on which problems are easy and difficult.
Most notably, \targetopsd{} and \method{} have a correlation of \(0.990\) after
pooling all 90 problems, and \method{} wins, ties, and loses on 31, 39, and
20 problems, respectively. Target-only and \method{} are similarly correlated
at \(0.989\). Replacing or removing the paired target solution, therefore, does
not substantially reorder problem-level difficulty in these runs. This
agreement does not establish a shared reasoning mechanism: common benchmark
difficulty can itself produce high correlations, and the pooled correlation
also contains between-benchmark variation.
We next bin problems by the observed Base count \(c_i^{(\mathrm{Base})}\).
Table~\ref{tab:base-difficulty-gains} reports the mean change
\(c_i^{(m)}-c_i^{(\mathrm{Base})}\) within each bin. The unit is additional
correct responses per problem out of 48, rather than percentage points.

\begin{table}[htbp]
\centering
\caption{\textbf{Accuracy-mass shifts by Base success count.}
Problems are binned by the number of correct Base responses out of 48.
Each method column gives the mean number of additional correct responses per
problem relative to Base.}
\label{tab:base-difficulty-gains}
\small
\setlength{\tabcolsep}{7pt}
\begin{tabular}{@{}lrrrr@{}}
\toprule
Base correct count & Problems & \targetopsd{} & \method{} & Target-only \\
\midrule
0      & 22 & \(+0.14\) & \(+0.14\) & \(+0.14\) \\
1--4   & 13 & \(+0.54\) & \(+0.85\) & \(+0.46\) \\
5--12  & 12 & \(+1.25\) & \(+0.92\) & \(+0.92\) \\
13--24 & 11 & \(+5.00\) & \(+5.09\) & \(+4.64\) \\
25--36 & 12 & \(+4.50\) & \(+7.08\) & \(+7.58\) \\
37--47 & 10 & \(+2.50\) & \(+2.50\) & \(+1.70\) \\
48     & 10 & \(-0.50\) & \(-0.70\) & \(-0.40\) \\
\midrule
All    & 90 & \(+1.71\) & \(+2.04\) & \(+1.94\) \\
\bottomrule
\end{tabular}
\end{table}

\begin{table}[htbp]
\centering
\caption{\textbf{Accuracy-mass shifts by Base success count for Qwen3-4B
non-thinking.}
Problems are binned by the number of correct Base responses out of 48.
Each method column gives the mean number of additional correct responses per
problem relative to Base.}
\label{tab:base-difficulty-gains-4b}
\small
\setlength{\tabcolsep}{8pt}
\begin{tabular}{@{}lrrr@{}}
\toprule
Base correct count & Problems & \targetopsd{} & \method{} \\
\midrule
0      & 45 & \(+1.13\) & \(+3.22\) \\
1--4   & 14 & \(+3.64\) & \(+2.07\) \\
5--12  & 10 & \(+4.10\) & \(+8.00\) \\
13--24 &  6 & \(-2.33\) & \(+5.50\) \\
25--36 &  3 & \(+9.33\) & \(+5.67\) \\
37--47 & 12 & \(+2.83\) & \(+1.42\) \\
48     &  0 & \textemdash & \textemdash \\
\midrule
All    & 90 & \(+2.12\) & \(+3.57\) \\
\bottomrule
\end{tabular}
\end{table}

All three distilled methods obtain their largest gains on problems for which
Base already succeeds at an intermediate rate. By contrast, each method adds
only three correct responses in total across the 22 Base-zero problems, or
\(0.14\) responses per problem.
The main empirical finding here is that the method increases the likelihood of producing a correct answer that was already accessible, rather than consistently uncovering solutions that Base sampling could not generate.
This descriptive analysis does not rule out individual newly solved problems.

\begin{table}[htbp]
\centering
\caption{\textbf{Problem-level agreement among Qwen3-4B non-thinking methods.}
Pearson correlations are computed from correct counts out of 48 responses per
problem. ``All'' pools the 90 problems from the three benchmarks. W/T/L
compares the first-named method with the second across those 90 problems.}
\label{tab:problem-level-agreement-4b}
\footnotesize
\setlength{\tabcolsep}{3.5pt}
\begin{tabularx}{\textwidth}{@{}Xrrrrc@{}}
\toprule
Pair (first vs.\ second) & AIME24 \(r\) & AIME25 \(r\) & HMMT25 \(r\) &
All \(r\) & W/T/L \\
\midrule
\targetopsd{} vs.\ Base
  & 0.938 & 0.975 & 0.928 & 0.948 & 42/36/12 \\
\method{} vs.\ Base
  & 0.838 & 0.831 & 0.966 & 0.870 & 43/34/13 \\
\method{} vs.\ \targetopsd{}
  & 0.883 & 0.857 & 0.922 & 0.882 & 37/29/24 \\
\bottomrule
\end{tabularx}
\end{table}

\section{Matched Thinking-Mode Control}
\label{sec:both-thinking-control}

The main Qwen3-1.7B experiments use a non-thinking student rollout and a
thinking teacher.  We test whether the target-solution \targetopsd{} gain
persists after removing this mode asymmetry.  In this control, both the
student rollout and the frozen teacher use thinking mode during training; the
training dataset, optimization budget, and distillation settings are otherwise
unchanged.  All conditions in Table~\ref{tab:both-thinking-control} are
evaluated in thinking mode on AIME 2024.

\begin{table}[htbp]
\centering
\caption{\textbf{Qwen3-1.7B matched thinking-mode control on AIME 2024.}
The both-thinking condition does not reproduce the Avg@12 gain observed with
the asymmetric training configuration.  Accuracy metrics are
four-seed means in percentages.  Avg@12 is reported with corrected Monte
Carlo standard error; Pass@12 and Vote@12 are reported with empirical standard error.}
\label{tab:both-thinking-control}
\small
\setlength{\tabcolsep}{5pt}
\resizebox{\textwidth}{!}{%
\begin{tabular}{@{}lccrrrr@{}}
\toprule
Condition & Student & Teacher & Avg@12 & Pass@12 & Vote@12 & Format \\
\midrule
Base
  & -- & --
  & \(49.86\pm0.94\)
  & \(76.67\pm0.00\)
  & \(70.00\pm2.72\)
  & 98.68 \\
\targetopsd{} 
  & Non-thinking & Thinking
  & \(\mathbf{55.42\pm0.82}\)
  & \(78.33\pm0.96\)
  & \(65.00\pm1.67\)
  & 99.17 \\
\targetopsd{} 
  & Thinking & Thinking
  & \(49.24\pm0.91\)
  & \(77.50\pm0.83\)
  & \(67.50\pm1.60\)
  & 85.90 \\
\bottomrule
\end{tabular}%
}
\end{table}

The both-thinking condition obtains \(49.24\) Avg@12, \(0.62\) percentage
points below Base and \(6.18\) points below the asymmetric \targetopsd{}
condition.  It therefore does not reproduce the main 1.7B improvement.
Pass@12 remains close to the other conditions, but per-sample accuracy and
answer-format validity are lower; in particular, format validity falls to
\(85.90\%\).  This result suggests that the observed 1.7B benefit is not
invariant to the student--teacher mode pairing.  Because each condition uses
one training seed, this control does not by itself identify whether the
difference is caused by mode asymmetry, rollout-state changes, or optimization
variance.

\junpei{slightly paraphrased}
\section{\method{} with physics References}
\label{sec:physics-donor-control}

Using Qwen3-4B in non-thinking mode, we test whether the teacher context remains effective when drawn from a non-mathematical domain.
We replace the teacher-only context pool with 2,768 problem–solution pairs sampled from the external CAMEL Physics dataset \citep{li2023camel}.
This matches the number of worked examples in the mathematics dataset. Both the student and teacher operate in non-thinking mode, and the two \method{} conditions are evaluated at the same checkpoint.

The following is a teacher prompt made during training.

\begin{tcblisting}{
  enhanced,
  breakable,
  colback=opnavy!2,
  colframe=opnavy!55,
  boxrule=0.7pt,
  arc=1.5mm,
  title={Actual teacher prompt with a physics reference},
  fonttitle=\bfseries,
  listing only,
  listing options={
    basicstyle=\ttfamily\scriptsize,
    breaklines=true,
    columns=fullflexible,
    keepspaces=true,
    showstringspaces=false
  }
}
Target mathematics problem:
Find $4^{-1} \pmod{35}$, as a residue modulo 35.
(Give an answer between 0 and 34, inclusive.)

Below is a worked example from a different and unrelated physics
problem. It is provided only to demonstrate careful scientific
reasoning and solution structure. It is not a solution or hint for the
target mathematics problem above. Do not reuse its final answer,
problem-specific quantities, units, laws, assumptions, or
domain-specific facts.

=== Unrelated physics Worked Example Problem Begin ===
A sample of radioactive material has an initial mass of 1 gram and a
half-life of 10 days. After 30 days, what will be the mass of the sample?
=== Unrelated physics Worked Example Problem End ===
=== Unrelated physics Worked Example Solution Begin ===
To find the mass of the radioactive material after 30 days, we can use
the formula:

Final mass = Initial mass * (1/2)^(time elapsed / half-life)

In this case, the initial mass is 1 gram, the half-life is 10 days, and
the time elapsed is 30 days. Plugging these values into the formula,
we get:

Final mass = 1 * (1/2)^(30 / 10)
Final mass = 1 * (1/2)^3
Final mass = 1 * (1/8)
Final mass = 0.125 grams

After 30 days, the mass of the radioactive sample will be 0.125 grams.
=== Unrelated physics Worked Example Solution End ===

Now solve the target mathematics problem independently. Use the worked
example only as a general demonstration of clear, step-by-step
problem-solving. Please reason step by step, verify your reasoning, and
put your final answer within \boxed{}.
\end{tcblisting}

\begin{table}[htbp]
\centering
\caption{\textbf{\method{} with physics problems.}
Replacing worked-problem mathematics solutions with physics solutions does not
retain the \method{} gain on any of the three benchmarks and substantially
reduces valid-answer formatting. }
\label{tab:physics-donor-control}
\small
\setlength{\tabcolsep}{4pt}
\resizebox{\textwidth}{!}{%
\begin{tabular}{@{}llrrrr@{}}
\toprule
Benchmark & Condition & Avg@12 & Pass@12 & Vote@12 & Format \\
\midrule
\multirow{3}{*}{AIME 2024}
& Base
  & \(23.19\pm0.72\)
  & \(50.83\pm1.60\)
  & \(33.33\pm2.36\)
  & 97.57 \\
& \method{} (mathematics)
  & \(\mathbf{31.53\pm0.80}\)
  & \(60.83\pm1.60\)
  & \(43.33\pm1.36\)
  & 89.58 \\
& \method{} (physics)
  & \(18.68\pm0.69\)
  & \(40.83\pm1.60\)
  & \(36.67\pm1.36\)
  & 46.04 \\
\midrule
\multirow{3}{*}{AIME 2025}
& Base
  & \(21.11\pm0.64\)
  & \(46.67\pm2.72\)
  & \(26.67\pm2.36\)
  & 99.17 \\
& \method{} (mathematics)
  & \(\mathbf{30.62\pm0.71}\)
  & \(55.00\pm2.15\)
  & \(40.83\pm1.60\)
  & 93.61 \\
& \method{} (physics)
  & \(20.21\pm0.61\)
  & \(44.17\pm2.10\)
  & \(29.17\pm2.50\)
  & 52.71 \\
\midrule
\multirow{3}{*}{HMMT 2025}
& Base
  & \(11.67\pm0.49\)
  & \(22.50\pm2.10\)
  & \(16.67\pm0.00\)
  & 99.58 \\
& \method{} (mathematics)
  & \(\mathbf{16.11\pm0.60}\)
  & \(38.33\pm2.15\)
  & \(20.00\pm1.92\)
  & 94.93 \\
& \method{} (physics)
  & \(8.68\pm0.50\)
  & \(20.83\pm2.10\)
  & \(15.00\pm0.96\)
  & 50.76 \\
\bottomrule
\end{tabular}%
}
\end{table}

The physics-reference condition obtains \(18.68\), \(20.21\), and \(8.68\)
Avg@12 on AIME 2024, AIME 2025, and HMMT 2025, respectively.  These values
are below Base by \(4.51\), \(0.90\), and \(2.99\) percentage points, and
below \method{} with mathematics references by \(12.85\), \(10.42\), and
\(7.43\) points.  Its Pass@12 is also below both comparison conditions on all
three benchmarks, while valid-answer formatting falls to approximately half
of the generations.  Thus, the failure to retain the mathematics reference
gain is consistent across the three evaluations rather than being specific
to AIME 2024.  In this configuration, an arbitrary worked solution from a
different scientific subject is not sufficient to reproduce the benefit of
the other-problem mathematical context.

This result does not isolate which property of the physics context causes the
decline.  The mathematics and physics pools differ in source, subject matter,
notation, units, solution style, context length, and subject-specific prompt
wording.  Moreover, the CAMEL solutions are synthetic and filtered but not
independently verified, and the external physics assignment is deterministic
but not pairwise aligned.
Accordingly, we treat this as a descriptive cross-subject ablation rather
than evidence that mathematical semantics alone explain \method{}.

\section{Exact Fixed Mathematical Contexts}
\label{sec:fixed-context-examples}

Table~\ref{tab:fixed-context-form-ablation} compares three versions of one
fixed mathematical context. All three conditions use the same auxiliary
problem. Within a condition, the displayed problem--solution pair is inserted
verbatim into the worked-problem fields of the \method{} teacher prompt and is
repeated across all training rows. This appendix reproduces the exact textual content of
the \texttt{problem} and \texttt{solution} fields used in those runs; line
wrapping below is typographic.

\subsection{Shared Fixed Problem}

\begin{tcblisting}{
  enhanced,
  breakable,
  colback=opmist,
  colframe=opnavy!70,
  boxrule=0.7pt,
  arc=1.5mm,
  title={Fixed auxiliary problem used in all three conditions},
  fonttitle=\bfseries,
  listing only,
  listing options={
    basicstyle=\ttfamily\scriptsize,
    breaklines=true,
    columns=fullflexible,
    keepspaces=true,
    showstringspaces=false
  }
}
Let $a, b, c$ be positive integers such that
$\frac{a}{77}+\frac{b}{91}+\frac{c}{143}=1$.
What is the smallest possible value of $a+b+c$?
\end{tcblisting}

\subsection{Concise Correct Solution}

This is the 204-token correct solution used by the \emph{Concise correct}
condition.

\begin{tcblisting}{
  enhanced,
  breakable,
  colback=opteal!3,
  colframe=opteal!75!black,
  boxrule=0.7pt,
  arc=1.5mm,
  title={Concise correct solution (204 tokens)},
  fonttitle=\bfseries,
  listing only,
  listing options={
    basicstyle=\ttfamily\scriptsize,
    breaklines=true,
    columns=fullflexible,
    keepspaces=true,
    showstringspaces=false
  }
}
Multiplying the equation by $1001=7\cdot 11\cdot 13$ gives
$13a+11b+7c=1001$. Let $S=a+b+c$. Then
$13(S-77)=2b+6c=2(b+3c)$. Because $13$ is coprime to $2$, the
positive integer $b+3c$ must be divisible by $13$, so $b+3c\ge 13$.
Hence $S-77=2(b+3c)/13\ge 2$, and therefore $S\ge 79$. Equality is
attained by $b=10$, $c=1$, and $a=68$: indeed,
$13\cdot68+11\cdot10+7\cdot1=1001$. Thus the smallest possible
value is $\boxed{79}$.
\end{tcblisting}

\subsection{Locally Corrupted Solution}

The corrupted condition keeps the same problem and nearly the same solution
form, but replaces the valid lower bound \(b+3c\ge 13\) with the false claim
that positivity requires the \emph{second} positive multiple of \(13\).
It consequently reports the feasible but nonminimal value \(81\).

\begin{tcblisting}{
  enhanced,
  breakable,
  colback=opred!3,
  colframe=opred!75,
  boxrule=0.7pt,
  arc=1.5mm,
  title={Locally corrupted solution (204 tokens)},
  fonttitle=\bfseries,
  listing only,
  listing options={
    basicstyle=\ttfamily\scriptsize,
    breaklines=true,
    columns=fullflexible,
    keepspaces=true,
    showstringspaces=false
  }
}
Multiplying the equation by $1001=7\cdot 11\cdot 13$ gives
$13a+11b+7c=1001$. Let $S=a+b+c$. Then
$13(S-77)=2b+6c=2(b+3c)$. Because $13$ is coprime to $2$,
positivity makes $b+3c$ at least the second positive multiple of
$13$, namely $26$. Hence $S-77=2(b+3c)/13\ge 4$, and therefore
$S\ge 81$. Equality is attained by $b=23$, $c=1$, and $a=57$.
This bound is sharp. Indeed,
$13\cdot57+11\cdot23+7\cdot1=1001$. Thus the smallest possible
value is $\boxed{81}$.
\end{tcblisting}

\subsection{Verbose Correct Solution}\label{sec:verbose}

The verbose condition retains the correct argument and answer, but expands the
derivation to 742 solution tokens.

\begin{tcblisting}{
  enhanced,
  breakable,
  colback=opgold!3,
  colframe=opgold!80!black,
  boxrule=0.7pt,
  arc=1.5mm,
  title={Verbose correct solution (742 tokens)},
  fonttitle=\bfseries,
  listing only,
  listing options={
    basicstyle=\ttfamily\scriptsize,
    breaklines=true,
    columns=fullflexible,
    keepspaces=true,
    showstringspaces=false
  }
}
Let
\[
S=a+b+c.
\]
We remove the fractions and translate the resulting equation into a
restriction on $S$. The least common multiple of $77$, $91$, and $143$
is $1001$. Multiplying the given equation by $1001$ gives
\[
13a+11b+7c=1001. \tag{1}
\]
Because $1001=13\cdot77$, we compare equation (1) with thirteen times
$S$. We have
\[
13S=13a+13b+13c.
\]
Subtracting equation (1) from this identity yields
\[
13S-1001=2b+6c=2(b+3c).
\]
Therefore
\[
13(S-77)=2(b+3c). \tag{2}
\]
Equation (2) implies that $13$ divides $2(b+3c)$. Since
$\gcd(2,13)=1$, it follows that
\[
13\mid(b+3c).
\]
Thus, for some integer $k$,
\[
b+3c=13k. \tag{3}
\]
The variables $b$ and $c$ are positive, so $b+3c$ is positive. It
follows that $k$ is a positive integer and in particular $k\ge1$.

Substituting (3) into (2), we get
\[
13(S-77)=2\cdot13k.
\]
Dividing by $13$ gives
\[
S=77+2k. \tag{4}
\]
Since $k\ge1$, equation (4) proves the universal lower bound
\[
S\ge77+2=79.
\]
It remains to show that equality is attainable. Equality requires
$k=1$, so equation (3) becomes
\[
b+3c=13.
\]
Take $c=1$. Then $b=10$, and equation (4) gives $S=79$. The
remaining variable is
\[
a=S-b-c=79-10-1=68,
\]
which is positive. Hence $(a,b,c)=(68,10,1)$ is an admissible
candidate if it satisfies the original constraint.

We can verify it first in the cleared-denominator equation:
\[
13\cdot68+11\cdot10+7\cdot1=884+110+7=1001.
\]
Equivalently, verification in the original equation gives
\[
\frac{68}{77}+\frac{10}{91}+\frac{1}{143}
=\frac{884}{1001}+\frac{110}{1001}+\frac{7}{1001}
=1.
\]
Thus this triple is valid and its sum is
\[
68+10+1=79.
\]
Equation (4) shows that every possible sum is $77+2k$ for some
positive integer $k$, so no valid triple has sum less than $79$. The
explicit triple above attains $79$. Therefore the smallest possible
value is
\[
\boxed{79}.
\]
\end{tcblisting}

\subsection{Trivial Correct Context}
\label{sec:fixed-context-trivial}

The trivial control replaces the fixed Omni-MATH example with the following arithmetic example. 

\begin{tcblisting}{
  enhanced,
  breakable,
  colback=opred!3,
  colframe=opred!75,
  boxrule=0.7pt,
  arc=1.5mm,
  title={Teacher prompt with the fixed trivial context},
  fonttitle=\bfseries,
  listing only,
  listing options={
    basicstyle=\ttfamily\scriptsize,
    breaklines=true,
    columns=fullflexible,
    keepspaces=true,
    showstringspaces=false
  }
}

Compute $1+1$.

Adding one and one gives two. Therefore, the final answer is \boxed{2}.

\end{tcblisting}

\section{Fixed-Context Replication with Qwen3-1.7B}
\label{sec:fixed-context-1b}

We repeat the fixed-context comparison with Qwen3-1.7B. 
The worked example here is the concise correct \texttt{omni02169} problem and solution pair (Section~\ref{sec:fixed-context-examples}), and the same pair is supplied to the teacher for every
target problem. The student rollout is collected in non-thinking mode, the
frozen teacher scores it in thinking mode, and evaluation uses the same
thinking-mode protocol as the main Qwen3-1.7B comparison. Table~\ref{tab:fixed-context-1b}
compares this condition with Base, target-solution \targetopsd{}, and the
standard \method{} condition that varies the worked-problem example across
training targets.

\begin{table}[htbp]
\centering
\caption{\textbf{Fixed mathematical context with Qwen3-1.7B.}
Repeating one concise, correct mathematical example matches the standard \method{} condition on AIME 2024 and HMMT 2025 and has a
higher Avg@12 point estimate on AIME 2025.
Within each benchmark, the highest Avg@12 point estimate is shown in
\textbf{bold} and the second-highest is \underline{underlined}.}
\label{tab:fixed-context-1b}
\begin{tabular}{@{}llrrr@{}}
\toprule
Benchmark & Condition & Avg@12 & Pass@12 & Vote@12 \\
\midrule
\multirow{4}{*}{AIME 2024}
& Base
  & \(49.86\pm0.94\) & \(76.67\pm0.00\) & \(70.00\pm2.72\) \\
& \targetopsd{}
  & \(\underline{55.42\pm0.82}\) & \(78.33\pm0.96\) & \(65.00\pm1.67\) \\
& \method{} 
  & \(55.35\pm0.83\) & \(75.83\pm0.83\) & \(70.00\pm2.36\) \\
& \method{} (fixed)
  & \(\mathbf{55.56\pm0.85}\) & \(76.67\pm1.36\) & \(70.00\pm1.36\) \\
\midrule
\multirow{4}{*}{AIME 2025}
& Base
  & \(37.36\pm0.81\) & \(70.00\pm3.04\) & \(48.33\pm0.96\) \\
& \targetopsd{}
  & \(40.35\pm0.77\) & \(65.00\pm2.15\) & \(50.00\pm1.36\) \\
& \method{} 
  & \(\underline{40.69\pm0.74}\) & \(63.33\pm2.36\) & \(52.50\pm1.60\) \\
& \method{} (fixed)
  & \(\mathbf{43.40\pm0.73}\) & \(65.83\pm1.60\) & \(53.33\pm2.36\) \\
\midrule
\multirow{4}{*}{HMMT 2025}
& Base
  & \(23.61\pm0.64\) & \(52.50\pm2.50\) & \(28.33\pm0.96\) \\
& \targetopsd{}
  & \(25.76\pm0.63\) & \(50.83\pm2.10\) & \(29.17\pm0.83\) \\
& \method{} 
  & \(\underline{27.57\pm0.66}\) & \(51.67\pm1.67\) & \(30.83\pm0.83\) \\
& \method{} (fixed)
  & \(\mathbf{27.78\pm0.66}\) & \(55.00\pm2.15\) & \(31.67\pm1.67\) \\
\bottomrule
\end{tabular}%
\end{table}

The fixed condition obtains $55.56$, $43.40$, and $27.78$ Avg@12 on
AIME 2024, AIME 2025, and HMMT 2025. Relative to varying-example \method{},
the corresponding point-estimate differences are $+0.21$, $+2.71$, and
$+0.21$ percentage points. It also matches or exceeds target-solution
\targetopsd{} on all three benchmarks. The result therefore, extends the
fixed-context observation beyond the Qwen3-4B non-thinking setting: exposure
to a diverse stream of worked examples is not necessary for obtaining an
OPSD-like gain in this Qwen3-1.7B run. It does not establish that a fixed
example is generally preferable. 

\section{Thinking-Mode Evaluation after Qwen3-4B Non-Thinking Training}
\label{sec:4b-nonthink-train-thinking-eval}

We test whether their behavior transfers across inference modes by evaluating
Base and the same trained models with thinking enabled, without further
training. 

\begin{table}[htbp]
\centering
\caption{\textbf{Thinking-mode evaluation after Qwen3-4B non-thinking
training.}
\method{} matches or exceeds \targetopsd{} in Avg@12 on all three
benchmarks, but neither distilled condition improves uniformly over the
thinking-mode Base model. Within each benchmark, the highest Avg@12 point
estimate is shown in \textbf{bold} and the second-highest is
\underline{underlined}; tied highest values are both bolded.}
\label{tab:4b-nonthink-train-thinking-eval}
\small
\setlength{\tabcolsep}{5pt}
\begin{tabular}{@{}llrrr@{}}
\toprule
Benchmark & Condition & Avg@12 & Pass@12 & Vote@12 \\
\midrule
\multirow{3}{*}{AIME 2024}
& Base
  & \(\mathbf{72.85\pm0.69}\) & \(86.67\pm0.00\) & \(80.83\pm0.83\) \\
& \targetopsd{}
  & \(\underline{70.90\pm0.73}\) & \(85.00\pm2.15\) & \(80.83\pm1.60\) \\
& \method{}
  & \(\mathbf{72.85\pm0.70}\) & \(88.33\pm1.67\) & \(79.17\pm0.83\) \\
\midrule
\multirow{3}{*}{AIME 2025}
& Base
  & \(\mathbf{67.36\pm0.73}\) & \(82.50\pm0.83\) & \(77.50\pm0.83\) \\
& \targetopsd{}
  & \(64.03\pm0.81\) & \(82.50\pm1.60\) & \(76.67\pm1.36\) \\
& \method{}
  & \(\underline{64.17\pm0.82}\) & \(83.33\pm1.36\) & \(78.33\pm0.96\) \\
\midrule
\multirow{3}{*}{HMMT 2025}
& Base
  & \(43.47\pm0.72\) & \(60.83\pm1.60\) & \(55.00\pm2.89\) \\
& \targetopsd{}
  & \(\underline{44.44\pm0.74}\) & \(68.33\pm2.15\) & \(55.83\pm0.83\) \\
& \method{}
  & \(\mathbf{45.69\pm0.71}\) & \(68.33\pm2.15\) & \(55.83\pm1.60\) \\
\bottomrule
\end{tabular}
\end{table}

Table~\ref{tab:4b-nonthink-train-thinking-eval} shows that the gains under non-thinking evaluation do not consistently transfer when
thinking is enabled at inference. This suggests that teacher supervision
should be produced in the same mode used for evaluation. Because both the
student rollout and the teacher were non-thinking during training, we interpret
this result as evidence for mode alignment rather than teacher-mode mismatch
alone.
\end{document}